\documentclass[lettersize,journal]{IEEEtran}

\ifCLASSINFOpdf
  \usepackage[pdftex]{graphicx}
\else
\fi

\usepackage{amsmath}
\usepackage{amssymb}
\usepackage{amsfonts}
\usepackage{bm}
\usepackage{color}
\usepackage{caption}
\usepackage{hyperref}
\usepackage{subcaption}
\hypersetup{
    colorlinks=true,
    linkcolor=blue,
    filecolor=blue,      
    urlcolor=blue,
    pdftitle={Overleaf Example},
    pdfpagemode=FullScreen,
    }
\usepackage{algpseudocode}
\usepackage{algorithm}
\usepackage{booktabs}
\usepackage{siunitx}
\usepackage{multirow}
\usepackage{orcidlink}

\begin{document}
%

\title{Coordinated Manipulation of Hybrid Deformable-Rigid Objects in Constrained Environments}

\author{Anees Peringal\orcidlink{https://orcid.org/0000-0002-7851-8542}, Anup Teejo Mathew\orcidlink{https://orcid.org/0000-0003-4022-7501}, Panagiotis Liatsis\orcidlink{https://orcid.org/0000-0002-5490-6030}, Federico Renda\orcidlink{https://orcid.org/0000-0002-1833-9809}

\thanks{Anees Peringal, Anup Teejo Mathew and Federico Renda are with the Department of Mechanical and Nuclear Engineering, Khalifa University, Abu Dhabi, UAE, and also with the Khalifa University Center for Autonomous Robotic Systems (KUCARS), Khalifa University, Abu Dhabi, UAE (e-mail:100045872@ku.ac.ae; anup.mathew@ku.ac.ae; federico.renda@ku.ac.ae)}
\thanks{Panos Liatsis is with the Department of Computer Science, Khalifa University, Abu Dhabi, UAE (e-mail:panos.liatsis@ku.ac.ae)}
        
}


\markboth{Journal of \LaTeX\ Class Files,~Vol.~14, No.~8, August~2015}%
{Shell \MakeLowercase{\textit{et al.}}: Bare Demo of IEEEtran.cls for IEEE Journals}

\maketitle
\begin{abstract}
Coordinated robotic manipulation of deformable linear objects (DLOs), such as ropes and cables, has been widely studied; however, handling hybrid assemblies composed of both deformable and rigid elements in constrained environments remains challenging. This work presents a quasi-static optimization-based manipulation planner that employs a strain-based Cosserat rod model, extending rigid-body formulations to hybrid deformable linear objects (hDLO). The proposed planner exploits the compliance of deformable links to maneuver through constraints while achieving task-space objectives for the object that are unreachable with rigid tools.
By leveraging a differentiable model with analytically derived gradients, the method achieves up to a 33× speedup over finite-difference baselines for inverse kinetostatic(IKS) problems. Furthermore, the subsequent trajectory optimization problem, warm-started using the IKS solution, is only practically realizable via analytical derivatives.
The proposed algorithm is validated in simulation on various hDLO systems and experimentally on a three-link hDLO manipulated in a constrained environment using a dual-arm robotic system. Experimental results confirm the planner’s accuracy, yielding an average deformation error of approximately 3 cm (5\% of the deformable link length) between the desired and measured marker positions. Finally, the proposed optimal planner is compared against a sampling-based feasibility planner adapted to the strain-based formulation. The results demonstrate the effectiveness and applicability of the proposed approach for robotic manipulation of hybrid assemblies in constrained environments.
\end{abstract}


\IEEEpeerreviewmaketitle

\section{Introduction}
Deformable linear objects (DLOs), such as cables, ropes, and wires, are ubiquitous in daily life and industrial applications. While recent research has focused heavily on the manipulation of DLOs \cite{tang_learning-based_2024,  sintov_motion_2020, bretl_quasi-static_2014,  yu_generalizable_2024, Naijing2022, Caporali2025}, many real-world objects are hybrid assemblies of deformable and rigid links. These systems leverage structural deformation to traverse highly constrained environments while retaining the mechanical precision inherent in conventional parallel designs. The continuum robotics community have proposed many such parallel continuum robots that have DLOs as well as rigid elements \cite{OrekhovParallelContinuum, BrysonParallel, Yan2021} in a Stuart-Gough fashion actuated by rotary motors that can be used for minimally invasive surgery. Similarly, coordinated aerial manipulation as in \cite{sun2025agile,de2025distributed} uses multiple drones to cooperatively transport a slung load through narrow gaps. In all of these cases, the underlying problem is the manipulation of assemblies of DLO and rigid elements. The differences between them arise from the morphology, i.e., the shape and connectivity of the different links, and from the actuation used. The following characteristics define the class of systems under investigation:
\begin{enumerate}
    \item Assembly of DLOs and rigid links.
    \item The requirement for coordinated actuation by two or more manipulators to achieve a target task-space objective.
    \item A closed-chain structure in which the object–manipulator system forms a parallel kinematic chain.
\end{enumerate}
We refer to such passive assemblies as hybrid Deformable Linear Objects (hDLOs), and to the corresponding robotic setups as hDLO–manipulator systems.

\begin{figure}[t]
    \centering
    \includegraphics[width=\linewidth]{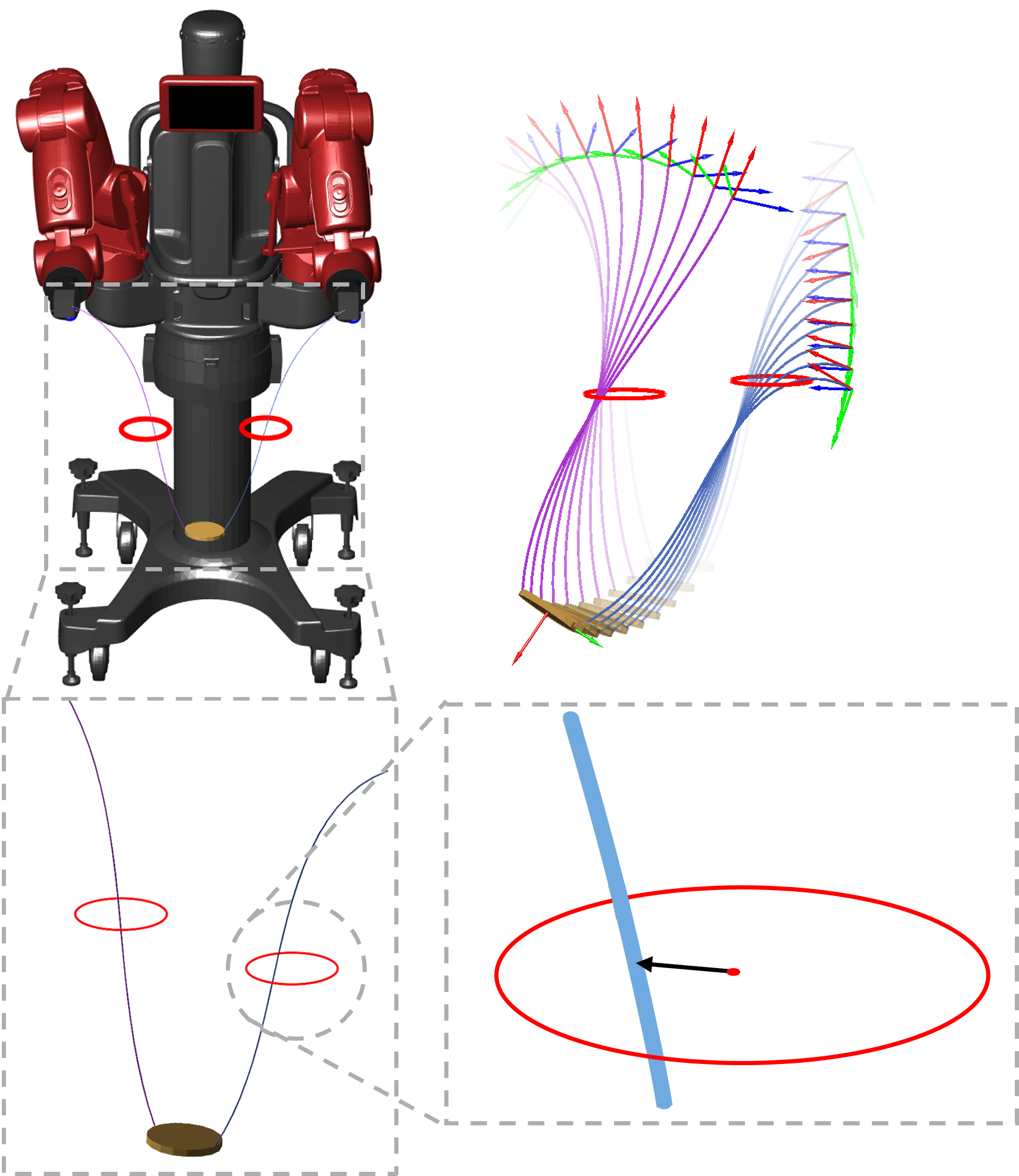}
    \caption{Overview of the robotic platform and an example assembly of deformable and rigid links. The goal of this paper is to generate a quasi-static trajectory that reaches a desired goal while satisfying the workspace (environmental) constraints represented by the red circles.}
    \label{fig1}
\end{figure}

The main challenges in the manipulation planning of hDLOs are inherited from their constituent DLOs, namely, a high-dimensional state-space and highly non-linear behaviour \cite{bretl_quasi-static_2014}. Because the shape of each DLO link in the assembly, i.e. the position and orientation of every point on the DLO link, is defined by a continuous mapping from the arc length to elements in $SE(3)$, the resulting configuration space of the hDLO is infinite dimensional. This infinite-dimensionality makes exhaustive or grid-based search strategies intractable. Consequently, sampling-based motion planning methods, such as rapidly exploring random trees (RRT) and probabilistic roadmaps (PRM), are commonly employed for DLO manipulation \cite{yu_generalizable_2024,bretl_quasi-static_2014}, as they are well-suited for efficiently exploring high-dimensional, complex configuration spaces \cite{lavalle2001randomized}. However, classical sampling-based planners do not guarantee convergence to an optimal trajectory within the feasible space. Manipulation of hDLOs necessitates a reduced-order model that consistently integrates DLO mechanics with rigid-body kinematics. This integration is naturally provided by the strain-based Geometric Variable Strain (GVS) formulation \cite{mathew_reduced_2025}. By combining rigid-body joint coordinates with rod strain parameters, the GVS model yields a finite-dimensional configuration space that is amenable to efficient motion planning. We demonstrate that the proposed GVS-based quasi-static planner computes optimal trajectories that steer the hDLO toward a target task-space objective while satisfying environmental constraints. Furthermore, we benchmark our approach against a sampling-based planner adapted for strain-based modelling of hDLOs.


Due to the passive nature of hDLO, the hDLO-manipulator system is underactuated. The planning challenge, therefore, lies in computing an optimal trajectory for the actuated Degrees of Freedom (DoFs) that drive the hDLO towards a target goal while satisfying environmental constraints. An example of this class of systems is shown in Fig. \ref{fig1}, where two manipulators coordinate to deform an hDLO, enabling it to traverse a constrained workspace. In this paper, we consider circular apertures of radius $\rho_h$ as an abstract model of narrow passages that commonly arise in applications like industrial inspection or minimally invasive surgery.

The key contributions of this work are:
\begin{itemize}
    \item We formulate a gradient-based quasi-static trajectory optimization framework for constrained manipulation of various hDLOs using a strain-based modelling approach.
    \item We provide a simulation study and experimental validation of the proposed method using a dual-arm robot manipulating an hDLO.
    \item We develop a sampling-based planner using the strain-based model and compare its performance with the proposed optimization framework.
\end{itemize}

The rest of the paper is organised as follows. Section \ref{Related_works} reviews the current state of the literature. Section \ref{GVS} summarises the GVS model applied to hDLOs in static equilibrium. Section \ref{Planning_alg} develops the optimization-based hDLO planner utilizing the GVS framework. Section \ref{Sim_results} applies the planning algorithm in simulation and explores various features of the planner. Section \ref{Exp_results} validates the planned trajectories on a real dual-arm robot by (i) verifying that the desired task-space objective is achieved and (ii) comparing experimental performance against simulation results. To evaluate the performance of our optimization-based planner against traditional sampling-based planners, we implement a Rapidly-exploring random tree (RRT) algorithm using the strain-based model in Section \ref{Comparison}. Finally, Section \ref{Conclusion} presents concluding remarks and directions for further exploration.

\section{Related works}\label{Related_works}
Effective planning of hDLOs requires models that balance computational efficiency with accurate representation of nonlinear deformations. We first review DLO modelling approaches, and then survey planning algorithms for deformable object manipulation, highlighting their respective strengths and limitations.
\subsection{hDLO Modeling}
Challenges in modelling hDLOs arise mainly from the modelling of the constituent DLOs and reconciling the rigid-body kinematics with the continuous deformation of the DLOs. Three-dimensional deformable objects can be modelled using the Finite Element Method (FEM), which discretizes the continuum equations governing material deformation into a numerically solvable form. FEM is used to model DLOs \cite{deng_robot-object_2024}, and can also be extended to rigid bodies and hDLOs that combine deformable and rigid components. However, FEM uses maximal coordinates, which can result in very high DoF systems. 

Another approach for modelling constituent DLOs is the discrete formulation, which assumes that the deformable object is composed of a finite set of interconnected elements. Within this class, lumped mass methods \cite{andrea_monguzzi_optimal_2025, Selle2008} have been used to model DLOs by representing them as point masses connected by compliant elements. In contrast, pseudo-rigid body models \cite{venkiteswaran2019shape} are particularly well suited for hDLOs, as the constituent DLOs are represented as rigid links connected by compliant joints, consistent with the rigid-joint modelling used in the overall assembly. These models represent hyperredundant robots well \cite{Khoshnam2013}, but are less suitable for continuously deforming objects due to their piecewise-rigid approximation.

Data-driven methods offer a bypass to analytical formulation by capturing the highly non-linear motion of deformable objects with fast computations \cite{sundaresan2020learning,huang2023learning}. While various neural-network architectures can be used, depending on the input data (e.g., visual, point cloud, etc), the resulting model is represented solely by the learned network parameters, making it difficult to interpret or analyse in the classical control-theoretic sense. 
In contrast, \cite{bruder1902modeling} proposes a Koopman-based method that produces a linear representation of the non-linear soft robot model in a lifted space. The linear system representation is well-suited for robotics since existing control methods can be directly applied to it. However, the lifted space may be very high-dimensional. Moreover, the inherent limitations of data-driven models in data requirements and generalizability also apply in this setting. 

The DLO links in the hDLO are naturally represented using rod models \cite{grinberg2025,bretl_quasi-static_2014, yu_generalizable_2024}. The Cosserat rod model provides a general six-mode deformation framework, from which many classical rod theories can be recovered as special cases. The Cosserat rod model can be solved by formulating it as a boundary value problem \cite{artinian_closed-loop_2024}. However, this solution can be unstable in some cases. Discrete elastic rods (DER) \cite{Bergou2008} are also used as a solution strategy that discretises the rod model into a finite number of vertices and edges consistent with continuum mechanics models. A major drawback is that this formulation of the Cosserat rod is more complex to apply to hDLOs because it does not naturally share the same joint abstraction. Another solution method uses the geometrically exact finite element method (GE-FEM) \cite{Simo1988}. The configuration space defined by this method is the field of cross-section poses of the DLO.
In contrast, the strain-based model used in this work represents the DLO through the strain along the DLO, which is analogous to the joint-space representation of hDLO. GVS is a general strain-based formulation of the Cosserat rod that allows for variable strains for the six modes of deformation. Other strain-based formulations, such as the widely used Piecewise Constant Curvature (PCC) \cite{Godage2016}, Piecewise Constant Strain (PCS) \cite{Renda2018}, can be seen as a special case of the GVS model.

\subsection{Planning for manipulation}
Planning algorithms can be broadly classified into sampling-based, optimization-based, and learning-based. Among these, sampling-based methods are widely used, as they approximate the configuration space by sampling feasible states and connecting them to form a graph that enables efficient navigation. A trajectory is then obtained by performing a graph search. Sampling-based approaches are particularly effective in high-dimensional configuration spaces, such as those encountered in hDLO manipulation. Notable examples of such methods include the Probabilistic Roadmap Planner (PRM) \cite{kavraki1996probabilistic} and the Rapidly-Exploring Random Tree (RRT) \cite{lavalle1998rapidly}. For a finite number of samples, the graph does not necessarily contain the globally optimal trajectory in configuration space; however, asymptotically optimal planners such as PRM* and RRT* converge to the optimal solution as the number of samples increases. A fundamental limitation of these methods is that the randomly sampled states may not inherently satisfy the hDLO's physical constraints, potentially leading to infeasible solutions. To address this, techniques such as projection onto the DLO equilibrium manifold \cite{yu_generalizable_2024} or sampling on the manifold \cite{bretl_quasi-static_2014}, have been proposed to ensure compliance with physical model. We develop a bidirectional RRT (BiRRT) tailored to strain-based modeling and conduct a simulation study to compare its performance with that of the proposed optimization framework.

Optimization-based methods are generally not used for manipulation planning of DLOs because they can be computationally expensive and are highly dependent on the initial guess. In \cite{aksoy2025planning}, the DLO is modelled with position-based dynamics (PBD) and a sampling-based planner is used to compute a trajectory as the initial guess for trajectory optimization. The good initial guess obtained from sampling makes trajectory optimization computationally feasible. This underscores the importance of proper initialisation. In contrast, we address this challenge through a dedicated warm-start strategy described in Sec. \ref{Sim_results}, without relying on a sampling-based planner. On the other hand, optimization has been widely used for local control of deformable objects. Model Predictive Control (MPC) is used along with data-driven models of DLOs, particularly in constrained environments, as it allows for the generation of control policies that adhere to environmental constraints. In \cite{tang_learning-based_2024}, the shape control problem of a DLO is addressed using MPC along with an offline data-driven DLO model.

Learning-based planners, including Reinforcement Learning (RL) and Imitation Learning (IL), find the policies directly from large-scale data about the system \cite{scheikl_sim--real_2023,han2017model}. Model-free reinforcement learning has been applied to DLO shape control via pick and place
\cite{wu2019learning}, though these methods remain sample-intensive and fail to generalise across varying material properties. To improve sample efficiency, IL leverages demonstrations to replicate human actions \cite{nair2017combining} or to define the problem \cite{li2023dexdeform}, and then refines the policy using differentiable simulation of the deformable object. While promising, their limitations in guaranteeing physical feasibility and adapting to different hybrid deformable–rigid assemblies motivate our model-based planning approaches with explicit physics. 

\section{Strain-based Modeling of hDLOs}\label{GVS}
\begin{figure}
    \centering           
    \includegraphics[width=0.90\linewidth]{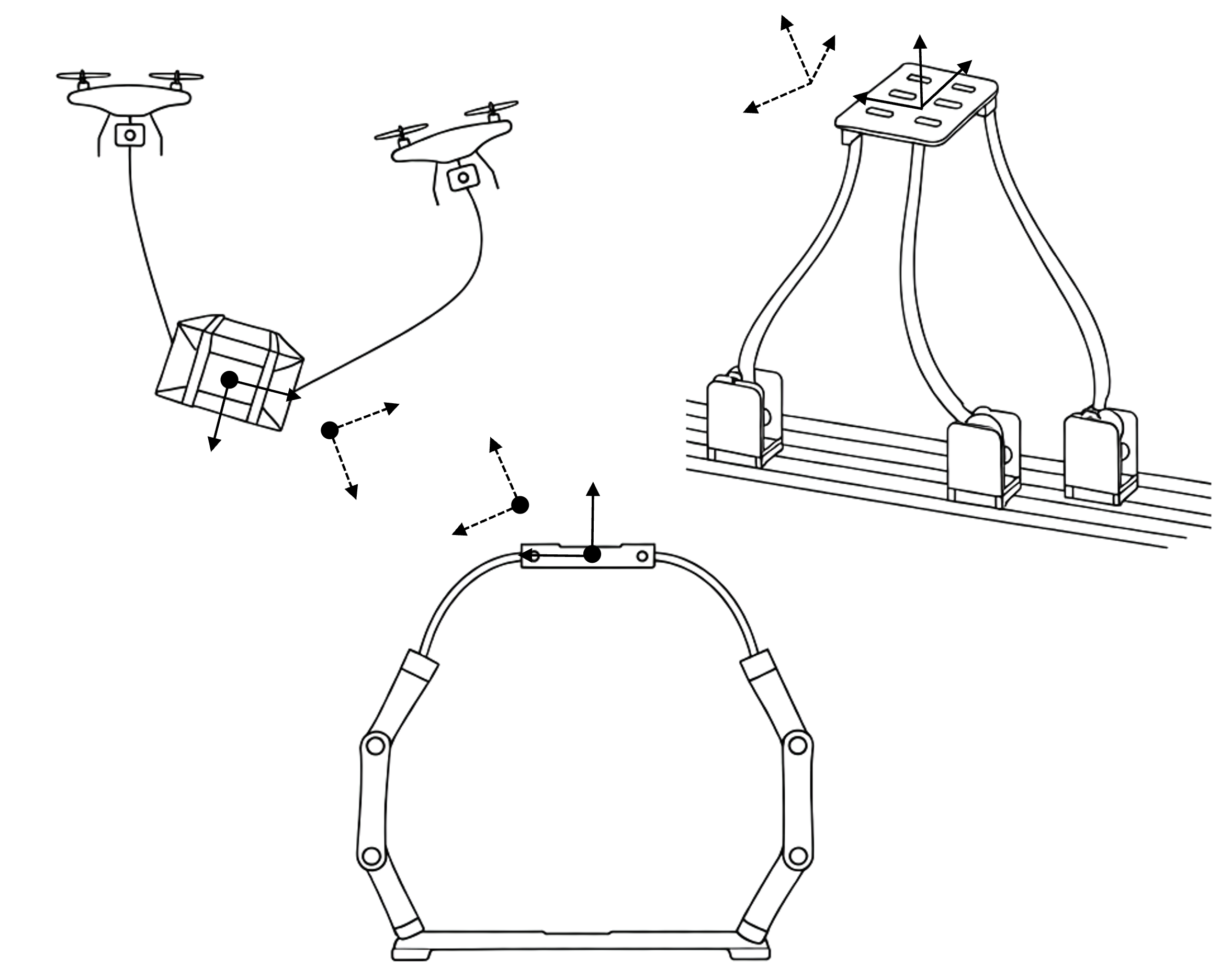}
    \caption{Schematics of exemplary hDLO systems. The depicted hDLO, composed of two DLOs, is actuated via aerial vehicles, prismatic and revolute joints, or serial manipulators with multiple revolute joints.}
    \label{fig:morphology_of_object}
\end{figure}

We adopt the Geometric Variable Strain (GVS) to model hDLO because it uses a minimal coordinate representation of the constituent DLOs \cite{mathew_reduced_2025}. Furthermore, GVS facilitates a unified treatment of rigid body joints and DLO deformation, allowing both to be expressed within the same geometrical framework. Schematics of some examples of hDLOs are shown in Fig. \ref{fig:morphology_of_object}.


The hDLO-manipulator system has $n_d$ total DoFs of which a subset $n_a$ are the actuated DoFs that parametrise the manipulator joint coordinates, while the rest represent the passive hDLO deformation. 
The homogeneous transformation matrix of the $i$th link with respect to the inertial base frame is given by: 
\begin{equation}
    \bm{g}_i = \begin{bmatrix}
        \bm{R}_i &\bm{r}_i\\
        \bm{0} &1
    \end{bmatrix} \in SE(3)
\end{equation}
where $\bm{r}_i\in \mathbb{R}^3$ and $\bm{R}_i \in SO(3)$ are the position and orientation respectively. 
In the case of a deformable link of length $l_i$, $\bm{g}_i$ is a function of the curvilinear abscissa $\bar{X} \in [0,l_i]$ and expressed with respect to the inertial base frame. We define a normalised abscissa $X = \bar{X}/{l_i}$. 

The partial derivatives of $\bm{g}_i(X)$ with respect to time and $X_i$ are given by:
\begin{align}\label{kinematics}
    \dot{\bm{g}_i}(X_i) = \bm{g}_i\hat{\bm{\eta}_i}\\
    \bm{g}_i'(X_i) = \bm{g}_i\hat{\bm{\xi}_i}\label{spacial_derivative}
\end{align}
where $\hat{\bm{\xi}_i}$ and $\hat{\bm{\eta}_i}$ are the strain and velocity twists represented in $\mathfrak{se}(3)$. By the equality of mixed partial derivatives in space and time, we obtain a relationship between the strain twist and the velocity twist:
\begin{equation}\label{mixed_derivatives}
    \bm{\eta}'
    _i= \dot{\bm{\xi
    }_i}-\mathrm{ad}_{\bm{\xi}_i}\bm{\eta}_i
\end{equation} 
The integration of \eqref{mixed_derivatives} yields:
\begin{equation}\label{velocity_twist}
    \bm{\eta}_i = \text{Ad}_{\bm{g}^{-1}_i}\int_0^{X_i}\text{Ad}_{\bm{g}_i}\dot{\bm{\xi}_i}ds
\end{equation}

The strain twist is, in general, a continuous function of $X_i$. For a link with $n_i$ DoF, the strain field is parametrised be a finite functional basis $\bm{\Phi} \in \mathbb{R}^{6\times n_i}$:
\begin{equation}\label{strain_parametrisation}
    \bm{\xi}_i = \bm{\Phi}_i(X_i)\bm{q}_i +\bm{\xi}_i^*(X)
\end{equation}
where $\bm{q}_i \in \mathbb{R}^{n_i}$ is the vector of generalized coordinates for the $i$th link, and $\bm{\xi}^* \in \mathbb{R}^6$ is the space rate of change of cross-section poses.
For rigid links, the basis vectors does not depend on $X_i$, and \eqref{strain_parametrisation} defines the joint twist. For example, in the case of a revolute joint about the z axis, we have $\bm{\Phi} = [0\; 0 \;1\; 0 \;0 \;0]^T$. This formulation can also handle multidimensional joints such as spherical joints and floating bodies. The forward kinematics for rigid joints can therefore be obtained by the integration of \eqref{spacial_derivative}: 
\begin{equation}\label{rigid_kinematics}
    \bm{g}_i = \exp(\hat{\bm{\xi}}_i)
\end{equation}In the case of a deformable link, the strain twist is generally a spatially varying function of $X_i$. To solve the differential equation \eqref{spacial_derivative}, we utilise the Magnus expansion \cite{Hairer2006}, which provides a matrix Lie algebra element $\bm{\hat{\Omega}}(X_i) \in \mathfrak{se}(3)$ such that the forward kinematics can be expressed through the exponential map:
\begin{equation}\label{fwdKinematics}
    \bm{g}_i(X_i) =\exp\left(\hat{\bm{\Omega}}_i(X_i)\right) 
\end{equation}
which effectively generalises rigid body kinematics in \eqref{rigid_kinematics}. In practice, $\bm{g}_i(X_i)$ is computed recursively at discrete points using Zanna quadrature \cite{zanna1999collocation} approximation of Magnus expansion.

The geometric Jacobian which maps $\bm{\dot{q}}$ to $\bm{\eta}_i$, $\bm{J}_i(\bm{q}) \in \mathbb{R}^{6\times n_d}$ and be obtained by substituting \eqref{strain_parametrisation}
in \eqref{velocity_twist}. where the first three rows, $\bm{J^\omega}$, correspond to the angular velocity and the last three rows, $\bm{J^v}$, describe the linear velocity. The generalised static equilibrium of the assembly is then obtained by projecting the freebody static equilibrium of each link $i$ using $\bm{J}_i$ to obtain \cite{Boyer2016}:
\begin{equation}\label{statics}
    \bm{Kq} -\bm{F(q)} - \bm{B(q)u} = \bm{0} 
\end{equation}
where $\bm{K}\in \mathbb{R}^{n_d\times n_d}$ is the generalized stiffness matrix, $\bm{B}\in\mathbb{R}^{n_d\times n_a}$ maps the actuator inputs to generlaized forces, $\bm{F} \in \mathbb{R}^{n_d}$ is the vector of external forces acting on the system such as gravity and $\bm{u}\in \mathbb{R}^{n_a}$ and actuation inputs such as joint torques of the manipulator. The vector $\bm{q}_a \in \mathbb{R}^{n_a}$ denotes the subset of generalized coordinates corresponding to the active DoFs.

The hDLO-manipulator systems considered in this paper have a parallel kinematic topology, which necessitates the satisfaction of one or more loop closure constraints. 
For a fully constrained closed-chain joint, between frames $A$ and $B$, the holonomic constraint represented in $SE(3)$ is given by:
\begin{equation}
    \bm{g}_A^{-1}(\bm{q})\bm{g}_B^{}(\bm{q}) = \bm{I}
\end{equation}
To facilitate numerical optimization, we map this constraint into the associated Lie algebra $\mathfrak{se}(3)$ through the logmap as:
\begin{equation}
    \bm{e}_c(\bm{q}) = \log(\bm{g}_A^{-1}\bm{g}_B) = \bm{0}
\end{equation}
where $\bm{e}_c \in \mathbb{R}^6$ represents the twist error associated with one fully constrained closed-chain joint. In general, the system may have multiple closed-chain joints resulting in a total $n_c$ scalar constraints: $\bm{e}_c \in \mathbb{R}^{n_c}$. The kinematic constraint in the Pfaffian form is given by $\bm{A}(\bm{q})\dot{\bm{q}} =\bm{0}$ \cite{armanini2021discrete}, where $\bm{A}$ is the constraint jacobian $\bm{A} = \frac{\partial \bm{e}_{c}}{\partial \bm q}$ (see Appendix \ref{appendix1}). The associated constraining forces lie in a space spanned by $\bm{A}^T$. Therefore, the constraining forces can be represented as $\bm{A}^T\bm{\lambda}$, where $\bm{\lambda} \in \mathbb{R}^{n_c}$ is the Lagrange multipliers that produce the force required to satisfy the constraint. Collecting all the closed chain constraints, our full static equilibrium condition will be:
\begin{subequations}
\label{statics}
\begin{align}
    \bm{Kq}- \bm{F} - \bm{Bu} -\bm{A^T\lambda} &=\bm{0}\label{statics1}\\
    \bm{e}_{c} &= \bm{0}\label{statics2}
\end{align}
\end{subequations}
In practice, $\bm{K}, \bm{F}$ and $\bm{A}$ for the continuous deformable object are computed through a numerical integration method. The continuous domain of $X$ is discretised into $n_p$ computation points, and the integral for these distributed quantities is computed as:
\begin{equation}
    \int_0^{1} \bm{f}(X)dX = \sum_{j=1}^{n_p} w_j\bm{f}(X_j)
\end{equation}
where $X_j$ and $w_j$ denote the integration points and their corresponding weights, respectively. The integrals are evaluated using Gauss–Legendre quadrature.

Recent advances provide closed-form analytical derivatives of the GVS statics and dynamics \cite{mathew_analytical_2025}, enabling efficient gradient-based optimization for hybrid soft–rigid robots. In this work, we focus on the partial derivatives of \eqref{statics} with respect to $\bm{q}$, $\bm{u}$, and $\bm{\lambda}$.

\section{Optimization based planning}\label{Planning_alg}
The challenge of quasi-static motion planning involves identifying an optimal path from an initial state to a goal state, constrained to the equilibrium manifold given by $\mathcal{M} = \{(\bm{q}, \bm{u},\bm{\lambda})\;|\;\mathrm{Eq.  }\; \eqref{statics} \}$. Since the systems under investigation are underactuated, the generalised coordinates cannot be independently prescribed. Thus, trajectories must be generated in the actuated subspace such that the resulting motion of the passive DoFs remains admissible. The admissibility is defined by constraints that depend on both passive and active DoFs such that the hDLO remains within a prescribed spatial region throughout the motion. We pose this trajectory generation as a constrained nonlinear optimization problem.

For hDLOs, the goal can be defined in terms of the position, orientation, or full pose of a designated link in the hDLO, referred to as the end-effector or a material point $X_i$ on a DLO link. Although the proposed framework accommodates all such goal specifications, in this work, we restrict our attention to either position, $\bm{r}_{EE} \in \mathbb{R}^3$, regulation or full pose, $\bm{g}_{EE} \in SE(3)$,  regulation of the hDLO end-effector, depending on the hDLO morphology. We first investigate the inverse kinetostatic problem. Subsequently, the solution is used to develop a warm-start strategy for the trajectory generation problem that satisfies the environmental constraints.

\subsection{Environment constraints}
The hDLO is constrained to remain within a spatial region throughout the motion. 
We model narrow passage environments using circular apertures that impose geometric constraints on the DLO links. Such confined geometries are commonly encountered in industrial inspection and minimally invasive surgery.
In this work, we consider two such circular apertures, and we assume that each aperture lies in a plane parallel to the $xy$-plane at height $z_h$. For each DLO link, there exists a normalised arc length $X^\dagger \in [0,1]$ where it intersects the aperture plane. For a single aperture, it can be written as a combination of equality and inequality constraints: 
\begin{equation}\label{hole_constraints}
\begin{split}
    c_{e}(\bm{q},X^\dagger)&:=z_h - z(\bm q, X^\dagger)=0\\ 
    c_{in}(\bm{q},X^\dagger )&:=\|\bm{P}_{h} - \bm{P}(\bm q,X^\dagger)\|^2 - (\rho_{h}-\rho(X^\dagger))^2\leq 0
    \end{split}
\end{equation}
where $\bm{P}_{h} = [x_{h}\;y_{h}]^T$ represents the $xy$-coordinates of the aperture center and $\bm{P}(\bm{q},X^\dagger) = [x(\bm{q},X^\dagger)\; y(\bm{q},X^\dagger)]^T$ is the $xy$-coordinates of the DLO link centerline at the point of intersection. $\rho_{h}$ and $\rho(X^\dagger)$ are the radius of an aperture and the cross-section of the DLO at the point of intersection, respectively. Since we consider two apertures, we will have two sets of $c_e$ and $c_{in}$ and the corresponding $\bm{X}^\dagger = [X^\dagger_1 \;X^\dagger_2]^T$. $\bm{X}^\dagger$ is treated as free variables within the optimization so that the hDLO can move along the aperture, while satisfying the constraints.

\begin{figure}
    \centering
    \includegraphics[width=0.8\linewidth]{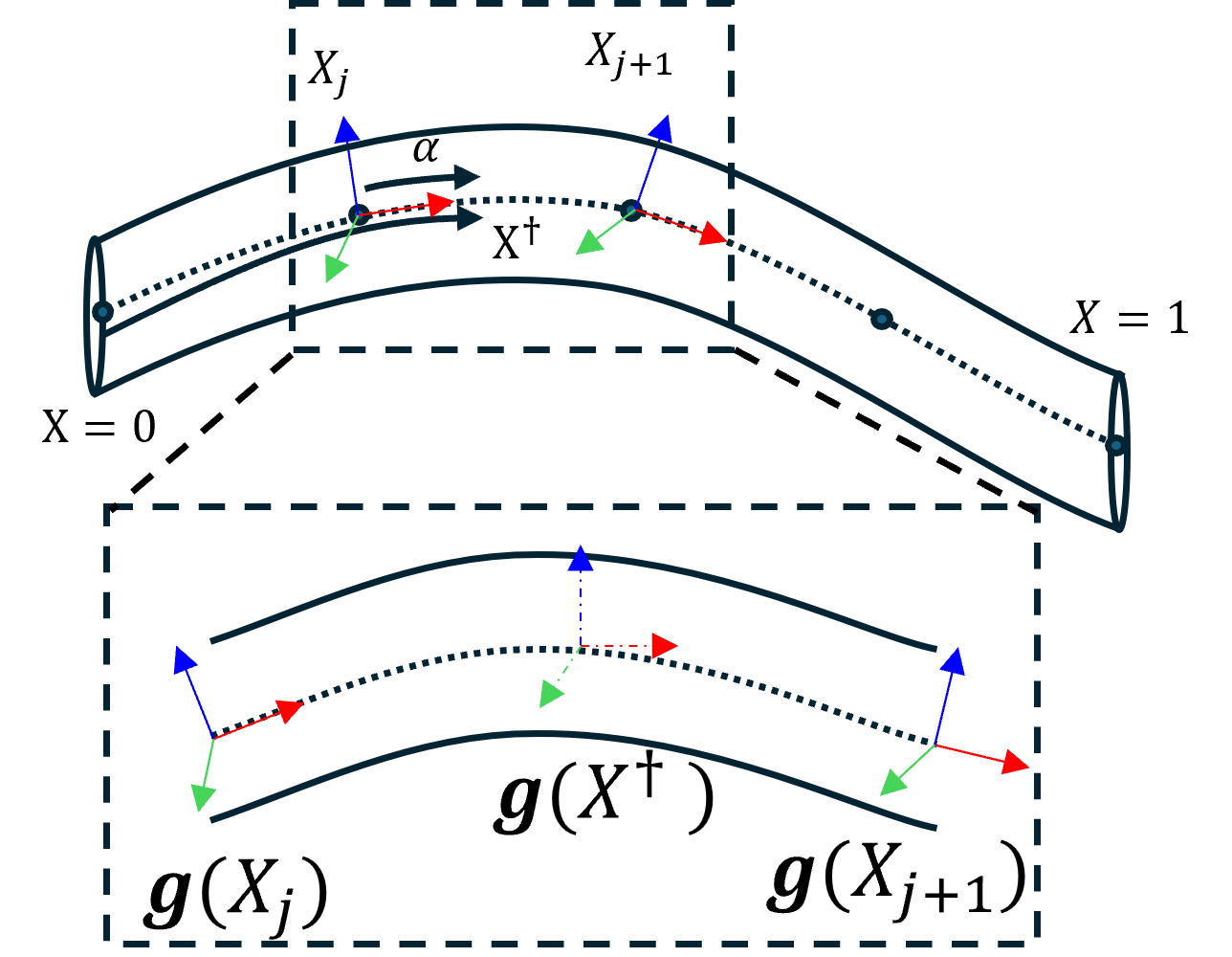}
    \caption{The pose of the cross section is computed at finite computational points denoted by $X_j$. At other points, we approximate the pose via an interpolation in $SE(3)$ by assuming a constant twist between two consecutive computational points. Interpolation is described in Algorithm \ref{Algorithm_interp}.}
    \label{fig :position_interpolation}
\end{figure}

The recursive algorithm of GVS framework provides the pose of the cross-section of the DLO links at discrete integration points as discussed in Sec \ref{GVS}. Hence, evaluating the environmental constraints requires the pose at arbitrary arc length $X^\dagger$. To achieve this, we utilise an interpolation scheme on $SE(3)$ to obtain a continuous representation of the DLO as illustrated in Fig. \ref{fig :position_interpolation}.

To evaluate the pose at an arbitrary arc length $X^\dagger$ that is between two known discrete points $X_j$ and $X_{j+1}$, we utilize interpolation as:
\begin{equation}
    \bm{g}(X^\dagger) = \bm{g}_j\bm{g}_\alpha\label{interpolation}
\end{equation}
where $\bm{g}_j=\bm{g}(X_j)$ is the pose at $X_j <X^\dagger$ and $\bm{g}_\alpha$ is given by $\exp\left(\alpha\bm{\Omega}_j\right)$, where $\alpha = \frac{X^\dagger - X_j}{X_{j+1}-X_{j}}$. This interpolation method is described in algorithm \ref{Algorithm_interp}.
\newcommand{\SO}{SO(3)}
\newcommand{\SE}{SE(3)}
\newcommand{\Exp}{\operatorname{Exp}}
\newcommand{\Log}{\operatorname{Log}}

\begin{algorithm}[H]
\caption{Zero-order Interpolation in \SE}\label{Algorithm_interp}
\begin{algorithmic}
\Require Keyframes $\{(X_j, \bm{g}_j)\}_{k=1}^{N_g}$ with $\bm{g}_j\in \SE$, and query coordinate $X^\dagger$
\Ensure Interpolated pose $\bm{g}(X^\dagger)\in \SE$
\State Find $j$ such that $X_j \le X^\dagger\le X_{j+1}$ 
\State $\alpha \gets \frac{X^\dagger - X_j}{\,X_{j+1} - X_j\,}$ 
\State $\hat{\bm{\Omega}}_j \gets \log(\bm{g}_j^{-1} \bm{g}_{j+1})$ 
\State $\bm{g}(X^\dagger) \gets \bm{g}_j \, \exp\big( \alpha\,\hat{\bm{\Omega}}_j \big)$ 
\State \Return $\bm{g}(X^\dagger)$
\end{algorithmic}
\end{algorithm}

To use gradient-based optimization, we require the analytical derivatives of the constraints with respect to the decision variables. The constraints in \eqref{hole_constraints} are defined by the DLO link position at $X^\dagger$, which is a function of $\bm{q}$ and $X^\dagger$. From the interpolated pose $\bm{g}(\bm{q},X^\dagger)$, the position vector is extracted as: 
\begin{equation}
    \bm{r}(\bm{q},X^\dagger) = \bm{R}_j\bm{r}_\alpha(\bm{q},X_j) +\bm{r}_j
\end{equation}
Using the chain rule, the analytical derivatives of the position of the deformable object with respect to $\bm{q}$ and $X^\dagger$ are given by:
\begin{equation}
\begin{split}
    \frac{\partial \bm{r}(\bm{q},X^\dagger)}{\partial \bm{q}} &= \frac{\partial \bm{R}_j}{\partial \bm{q}}\bm{r}_\alpha +\bm{R}_j\frac{\partial \bm{r}_\alpha}{\partial \bm{q}} +\frac{\partial \bm{r}_j}{\partial \bm{q}}\\
    \frac{\partial\bm{r}(\bm{q},X^\dagger)}{\partial X^\dagger} &= \bm{R}_j\frac{\partial \bm{r}_\alpha}{\partial X^\dagger} +\bm{r}_j
\end{split}
\end{equation}
The full derivation of these terms and the derivatives with respect to $\bm{q}$ and $\bm{X}^\dagger$ are provided in Appendix \ref{Ap:postion_derivatives}. We use this definition of environmental constraints to define the inverse kinetostatic problem and, subsequently, the trajectory-generation problem.

\subsection{Inverse kinetostatic problem}
Inverse Kineto-Statics (IKS) for hDLOs consists of determining an equilibrium configuration and the corresponding actuation inputs required to achieve a desired kinematic goal. Unlike classical inverse kinematics, the configuration is implicitly determined by the static equilibrium of the hDLO-manipulator system. This problem is formulated as a nonlinear constrained optimization problem that minimizes the objective (goal) function. For example, when the goal is full pose regulation, the objective can be defined using the $SE(3)$ logarithmic distance between the desired pose $\bm{g}_{EE,d}$ and the end-effector pose $\bm{g}_{EE}(\bm{q})$:
\begin{equation}\label{InvKinStatic}
\begin{split}
    \min_{\bm{x}} \quad \frac{1}{2}&\|\mathrm{log}\left(\bm{g}_{EE,d}^{-1}\bm{g}_{EE}(\bm{q})\right)^\vee\|^2\\
    \mathrm{s.t.}\quad& \bm{Kq} -\bm{Bu} - \bm{F} - \bm{A}^T\bm{\lambda} = \bm{0}\\
    & \bm e_{c}(\bm{q}) = \bm 0\\
    & \bm c_{eq}(\bm{q},\bm{X}^\dagger)=\bm 0\\ 
    & \bm c_{in}(\bm{q},\bm{X}^\dagger)\leq \bm 0\\
    & \bm{q}_{a,\text{min}}\leq \bm{q}_a \leq \bm{q}_{a,\text{max}}
    \end{split}
\end{equation}
where $\bm{x} = [\bm{q}^T\;\bm{u}^T\; \bm{\lambda}^T \; \bm{X}^\dagger ]^T$ and $\bm{q}_{a,\text{min}}$ and $\bm{q}_{a,\text{max}}$ are the joint limits of the manipulator.
We solve the inverse kinetostatic problem using the \textit{fmincon} solver in MATLAB with the interior-point algorithm with analytical derivatives specified for the objective function and the constraints. These analytical derivatives are given in Appendix \ref{Ap:InvKinematics_derivatives}. The known initial configuration of the hDLO, obtained from static equilibrium \eqref{statics}, is used as the initial guess for the solver. The solution of \eqref{InvKinStatic} is subsequently used to obtain a warm start for the path planning problem. 

\subsection{Path planning problem}
The path planning problem is formulated as finding an optimal trajectory in the configuration space that transitions from the initial configuration to a final configuration that achieves a desired hDLO end-effector pose. 
The planner computes a sequence of actuation inputs and hDLO configurations that move the hDLO from start to goal while satisfying the constraints. We formulate the trajectory optimization (TO) as a non-linear program, where the decision variables $\bm{x}$ are parameterized along a path parameter $t$ and defined as $\bm{x}(t) = [\bm{q}(t)^T, \bm{u}(t)^T, \bm{\lambda}(t)^T, \bm{X}^\dagger(t)]^T$. To solve this problem numerically, we employ direct transcription method \cite{pardo_evaluating_2016} to transform the continuous functions into $N$ discrete keyframes such that $\bm{x}_k = [\bm{q}_k^T, \bm{u}_k^T, \bm{\lambda}_k^T, \bm{X}^{\dagger T}_k]^T$ for $k \in \{1,\dots, N\}$. We assume that the initial configuration $[\bm{q_0}, \bm{u_0}, \bm{\lambda_0}]$ is known a \textit{priori}. The optimization problem consists of $N\times (n_d +n_a +n_c +2)$ decision variables accounting for $\bm{q}_k$, $\bm{u}_k$, $\bm{\lambda}_k$ and $\bm{X}^{\dagger}_k$ respectievely. We formulate the TO problem as: 
\begin{equation}\label{traj_optimization}
\begin{split}
    \min_{\bm{x}} \quad &f_{p}(\bm{x})\\
    \mathrm{s.t.}\quad&\bm{K q}_k-\bm{B}_k\bm{u}_k - \bm{F}_k - \bm{A}_k^T\bm{\lambda}_k =\bm{0}\\
    &\bm{e}_{c}(\bm{q}_k)= \bm{0}\\
    &\bm c_{eq}(\bm{q}_k,\bm{X}^\dagger_k)= \bm 0\\ 
    &\bm c_{in}(\bm{q}_k,\bm{X}^\dagger_k)\leq \bm 0\\
    & \bm{q}_{a,\text{min}}\leq \bm{q}_{a_k} \leq \bm{q}_{a,\text{max}} \quad \forall\; k \in \{1,\dots, N\}\\
    &\frac{1}{2}\|\mathrm{log}\left(\bm{g}_{EE,d}^{-1}\bm{g}_{EE}(\bm{q}_N)\right)^\vee\|^2 = 0
    \end{split}
\end{equation}
where $f_p$ is the path cost of the optimization problem. The path cost promotes smooth variations in deformation, actuation inputs, and constraint forces by penalizing the differences between consecutive keyframes. It is defined as a weighted quadratic norm:

\begin{align}
    f_p &= \sum_{k=0}^{N-1}\|\bm{q}_{k+1}-\bm{q}_k\|^2_{Q_q}+\|\bm{u}_{k+1}-\bm{u}_k\|^2_{Q_u}+\label{terminal_cost_1}\\&\qquad\|\bm{\lambda}_{k+1}-\bm{\lambda}_k\|^2_{Q_\lambda}\notag
\end{align}
This creates a large NLP with $N\times (n_d+n_c+4)+1$ nonlinear constraints from the static equilibrium $N\times (n_d+n_c)$, environmental constraint $N\times4$, and a terminal constraint. The terminal constraint enforces the desired hDLO end-effector pose at the final keyframe. Moreover, the Jacobians are sparse because the keyframes in the trajectory are independent. We utilise the \textit{fmincon} solver with the \textit{interior-point} algorithm to solve the optimization problem, as it is particularly suited to large-scale sparse problems. 
To avoid the slow convergence associated with a cold-start initialization, $\bm{x}_k = \bm{x}_0$, we employ a warm-start strategy to improve efficiency. First, we compute the IKS solution of \eqref{InvKinStatic} to obtain a feasible configuration of the hDLO that achieves the desired pose. This provides the final state $\bm{x}_f$. The initial guess for the intermediate keyframes is then obtained by linearly interpolating between the initial state $\bm{x}_0$ and the feasible final state $\bm{x}_f$:

\begin{equation}
\label{warm_start_interpolation}
\bm{x}_k = \bm{x}_0\left(1-\frac{k}{N}\right) + \bm{x}_f\frac{k}{N}.
\end{equation}
This warm-start initialization improves the convergence of the TO. In contrast, cold-start initialization often results in slow convergence or failure to converge for several desired hDLO end-effector poses tested in simulation. Moreover, we provide the analytical Jacobians of the objective and constraint functions to the solver. The explicit expressions for these derivatives are given in the Appendix \ref{Ap:path_planning_derivatives}.

\section{Simulation results}\label{Sim_results}

\begin{figure*}
    \centering
    \includegraphics[width=\linewidth]{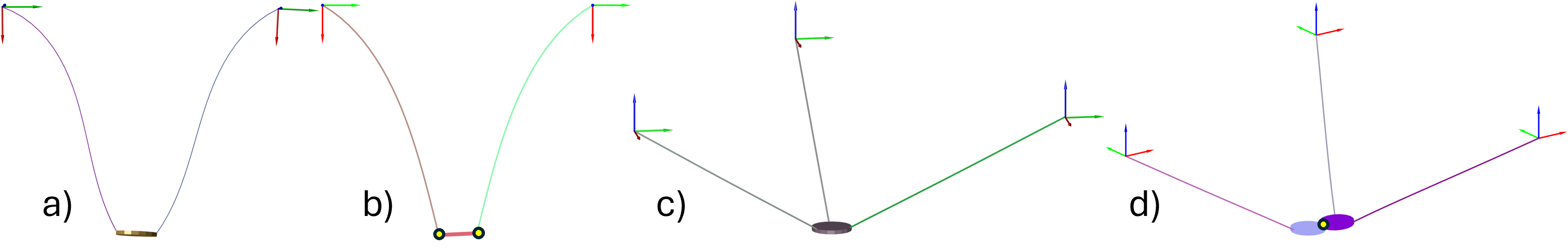}
    \caption{Deformable-rigid assembly examples: a) Deformable and rigid links, connected by a fixed link. b) Deformable rigid links, connected by spherical joints represented by a yellow circle. c) Two deformable and single rigid links d) Three deformable and two rigid links, connected by spherical joints.}
    \label{fig:sim_examples}
\end{figure*}

The SoRoSim simulator \cite{mathew_sorosim_2023} is used to implement the GVS model of the hDLO. We apply the proposed method to various hDLO assemblies of increasing complexity, as shown in Fig.~\ref{fig:sim_examples}. hDLOs (a) and (b) consist of two DLO links connected to a rigid link. In the first case, the connections between the DLO links and the rigid link are fixed, whereas in the second case spherical joints are used. The hDLO in (c) has three DLO links connected to a rigid link through fixed joints. Finally, hDLO (d) also consists of three DLO links connected to two rigid links via fixed joints, while the rigid links are connected to each other through spherical joints. In this case, one of the rigid links is selected as the hDLO end-effector, whose pose must achieve the desired goal.

We consider hDLOs (a) and (b) to be actuated by a dual-arm manipulator, where the pose is determined using the product-of-exponentials representation of the manipulator joint coordinates. For hDLOs (c) and (d), the DLO tip pose is modeled using exponential coordinates in $\mathbb{R}^6$. For hDLOs (b) and (d), which contain spherical joints, the control objective is restricted to end-effector position because spherical joints do not transmit arbitrary moments and therefore cannot fully constrain orientation.

The material properties of the DLO links and the rigid links used to make up the hDLO are given in Table \ref{tab:Nitinol_parameters}. 
The strain field along the DLO links is parametrised using a finite-order functional basis. In this work, we use third-order Legendre polynomials to represent all six deformation modes of the Cosserat rod. Hence, in total for a DLO link $n_{i} = 6\times4 = 24$. 

\begin{table}[t]
\centering
\caption{Geometric and material properties of the assembly components used in the simulation.}
\label{tab:Nitinol_parameters}
\begin{tabular}{lll}
\toprule
\textbf{Component} & \textbf{Parameter} & \textbf{Value} \\
\midrule
\multirow{6}{*}{DLO (Nitinol)} 
& Length & \SI{0.68}{\meter} \\
& Outer diameter & \SI{1.8}{\milli\meter} \\
& Inner diameter & \SI{1.4}{\milli\meter} \\
& Young's modulus & \SI{7.5e10}{\pascal} \\
& Poisson ratio & 0.33 \\
& Density & \SI{6450}{\kilogram\per\meter\cubed} \\
\midrule
\multirow{3}{*}{Rigid disk} 
& Length & \SI{0.01}{\meter} \\
& Diameter & \SI{0.1}{\meter} \\
& Mass & \SI{48}{\gram} \\
\bottomrule
\end{tabular}
\end{table}

\subsection{Simulation Results}
\begin{figure}
    \centering
    \includegraphics[width=\linewidth]{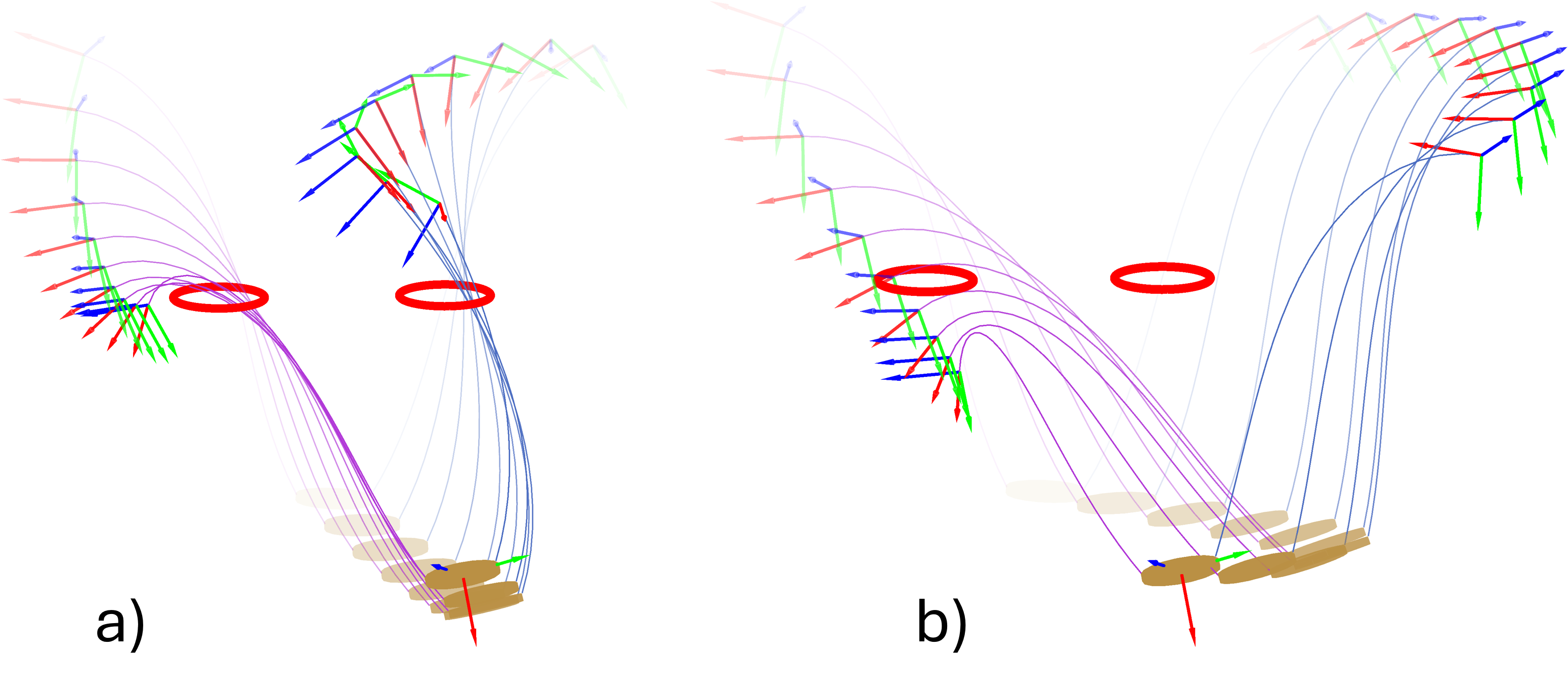}
    \caption{Moving from the same initial configuration of the robot, we need to make use of the deformation of the hDLO to move to the desired pose within the hole constraints. a) shows the trajectory when constraints are considered. b) shows the trajectory when the constraints are not considered.}
    \label{fig:constrained_vs_unconstrained.png}
\end{figure}

Using the hDLO in Fig.~\ref{fig:sim_examples}(a), we investigate the effect of environmental constraints. The NLP \eqref{traj_optimization} is solved with and without environmental constraints. In both cases, the terminal pose constraint 
$\frac{1}{2}\|\mathrm{log}(\bm{g}_{EE,d}^{-1}\bm{g}_{EE}(\bm{q}_N))^\vee\|^2 = 0$
is satisfied within numerical precision. When environmental constraints are introduced, the resulting trajectory differs qualitatively from the unconstrained case: the planner actively exploits the deformation of the hDLO links to satisfy the environmental restrictions, as illustrated in Fig.~\ref{fig:constrained_vs_unconstrained.png}. The constraints reshape the feasible configuration space, leading the optimizer to a different equilibrium configuration that accommodates both the task and environmental constraints.

\begin{figure*}
     \centering
     \includegraphics[width =\textwidth]{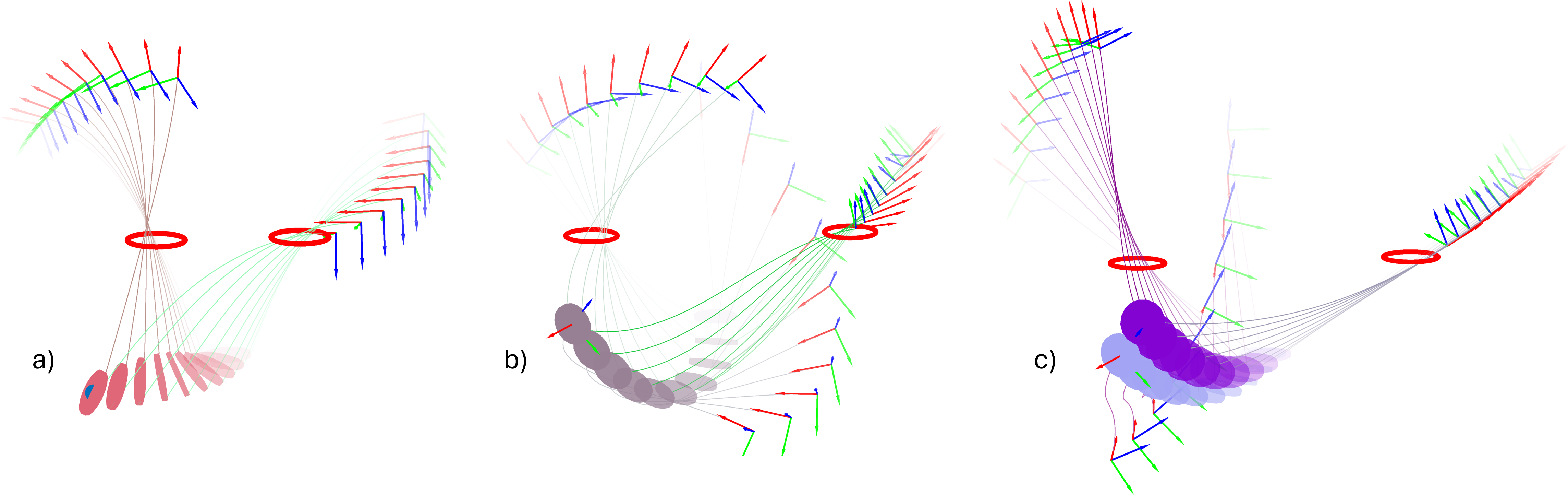}
        \caption{TO planner is extended to other hDLOs in Fig \ref{fig:sim_examples}(b), (c) and (d) respectively. The desired position for hDLO (b) is shown by the blue point and similarly the desired poses for other hDLOs are shown with the axes.}
        \label{fig:Sim examples}
\end{figure*}

We extend the analysis of the constrained NLP to the remaining hDLOs shown in Fig.~\ref{fig:sim_examples}. The resulting trajectories are presented in Fig.~\ref{fig:Sim examples}. In all cases, the terminal constraint is satisfied within numerical precision. The corresponding trajectory computation times and solver iterations are reported in Table~\ref{tab:assembly_time_taken}. The variation in computation time across assemblies is influenced by the number of deformation variables $n_d$, the morphology of the hDLO, and the terminal constraint (goal). For instance, moving from assembly (a) to (b), the convergence time nearly doubles despite the addition of only 6 DoFs. This increase is primarily due to the presence of passive spherical joints in the hDLO, which introduce additional kinematic coupling that the optimizer must resolve.

Overall, the results demonstrate that the proposed framework scales reasonably with problem size and remains computationally tractable even in the presence of environmental constraints.

\subsection{Effect of Analytical Derivatives}
The IKS \eqref{InvKinStatic} and the TO \eqref{traj_optimization} are solved using \textit{fmincon} with analytical derivatives supplied to the solver. When derivatives are not provided, the solver approximates them using finite differences. We evaluate the benefit of supplying analytical derivatives by comparing the two approaches on the hDLO assembly shown in Fig.~\ref{fig:sim_examples}(a).

Finite differences result in an average convergence time of 495.17\,s, whereas the use of analytical derivatives reduces this to 14.71\,s, corresponding to a $33\times$ speedup for the IKS problem. The average number of solver iterations also decreases from $373$ to $282$, indicating improved convergence behavior when analytical derivatives are provided. 

For the TO, a quantitative comparison is not possible because the solver did not converge within a reasonable time when gradients were approximated using finite differences. In contrast, when analytical derivatives are supplied to the solver, the TO converges reliably. This indicates that providing analytical derivatives is essential for solving the trajectory generation problem within a practical computation time.

\begin{table}[]
    \centering
    \begin{tabular}{c c c c c c}
         \hline Assembly & $n_d$& $n_a$& $n_c$ & Time taken (s) & Iteration count\\\hline
         a & 62 & 14 & 6& 27.66 & 221\\
         b & 68 & 14 &6 &52.88& 141\\
         c & 90 & 18 & 12 &53.38& 132\\
         d & 93 & 18 & 12 &55.18&  102\\\hline
         
    \end{tabular}
    \caption{The time taken to generate the trajectories shown in  Fig. \ref{fig:constrained_vs_unconstrained.png}(a) and Fig. \ref{fig:Sim examples}.}
    \label{tab:assembly_time_taken}
\end{table}

\section{Experimental results}\label{Exp_results}
To validate the proposed planning framework, we conducted experiments using a bimanual robot manipulating the hDLO shown in Fig.~\ref{fig:sim_examples}(a). The hDLO is constructed from highly elastic Nitinol rods, while the rigid link is 3D printed using polylactic acid (PLA). The geometric and material properties of the Nitinol and rigid segments are listed in Table~\ref{tab:Nitinol_parameters}.

To ensure accurate tracking of the hDLO, we perform hand–eye calibration using an \textit{Optitrack} motion capture system. Motion capture markers are attached to the manipulator end-effectors. The calibration procedure aligns the robot base frame with the global motion capture frame and accounts for the fixed geometric offset between the manipulator end-effector and the hDLO grasp points. This unified kinematic chain, which includes the \textit{Optitrack} base frame, manipulator, and the hDLO, allows the system to track both the hDLO configuration and the end-effector pose during execution. During the experiments, additional motion capture markers are placed on the hDLO end-effector to compare the measured end-effector pose with the desired pose.


\subsection{Forward Model Validation}
The accuracy of the hDLO–manipulator model was validated by comparing experimental measurements of the hDLO end-effector pose with the model's forward statics solution. For a set of $N_v$ arbitrary manipulator joint configurations, we computed the equilibrium state of the hDLO in simulation and compared the resulting end-effector poses with those measured using the motion capture system. The pose is compared using the positions of four markers in simulation and experiment through the following error function:
\begin{equation}\label{error_function}
e_m = \frac{1}{N_v}\sum_{k=1}^{N_v}\left(\frac{1}{4}\sum_{l=1}^{4}\|\bm{p}_{l}^{\,\text{sim}}(\bm{q}_k^*)-\bm{p}_{k,l}^{\,\text{exp}}\|\right)
\end{equation}
where $\bm{p}_{l}^{\,\text{sim}}(\bm{q}_k^*)$ is the simulated position of marker $l$ obtained from the forward statics solution at configuration $\bm{q}_k^*$, and $\bm{p}_{k,l}^{\,\text{exp}}$ is the corresponding experimental measurement from the motion capture system.

\begin{figure}
    \centering
    \includegraphics[width=
    \linewidth]{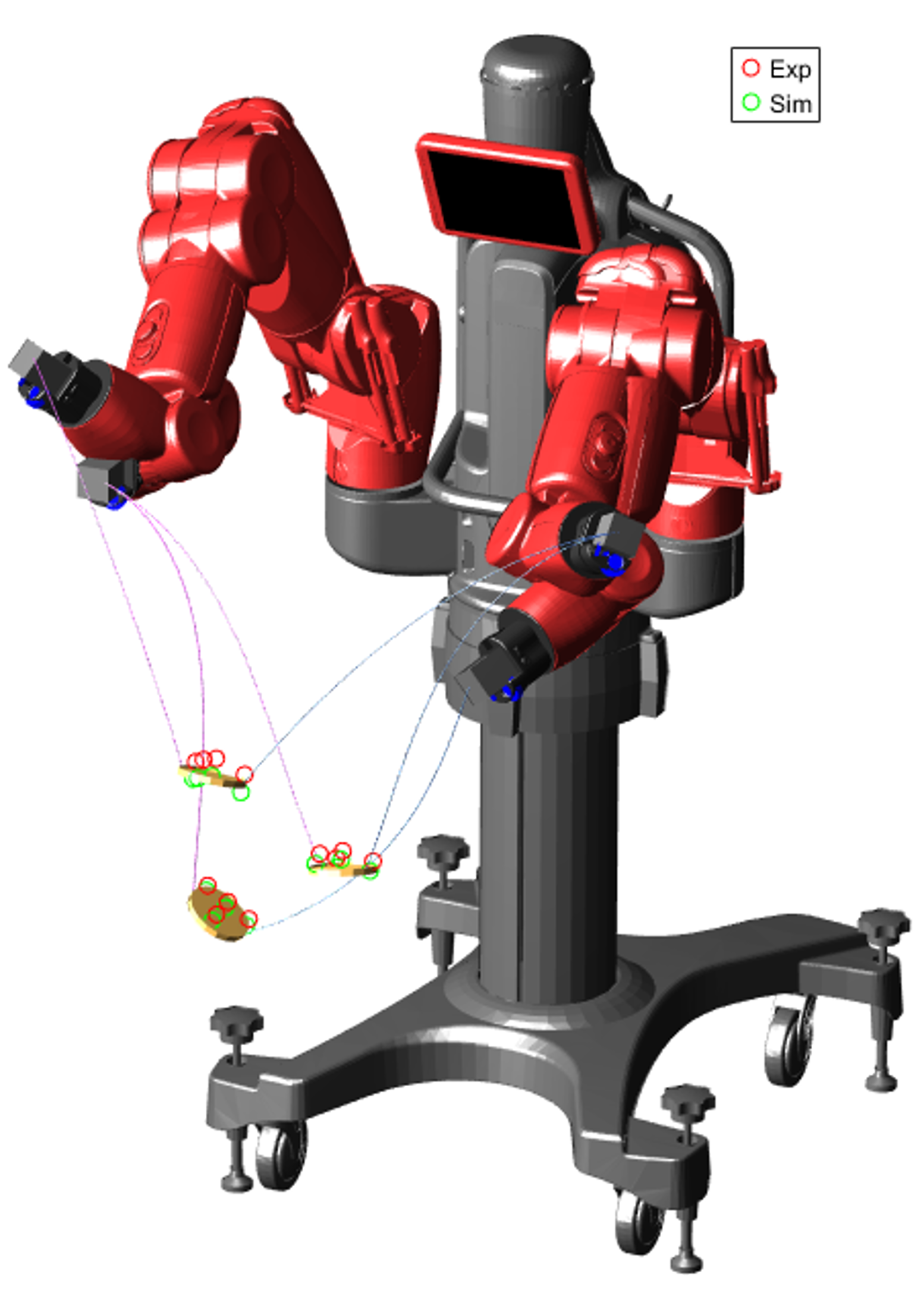}
    \caption{Experimental validation of the hDLO-manipulator model. The average error between the markers in experiment and simulation is 3.07\% of the deformable object length.}
    \label{fig:model_validation}
\end{figure}

As shown in Fig. \ref{fig:model_validation}, the model accurately predicts the steady-state configuration of the hDLO across the workspace. The validation results yielded an average error of 2.14 cm, which is 3.07\% of the length of one of the deformable links. 

\subsection{Manipulation experiments}
We validate the TO framework through four distinct experimental cases, each targeting a different desired hDLO end-effector pose. For each case, the optimal trajectory was computed both without and with environmental constraints.

To translate the discrete keyframe joint configurations obtained from solving \eqref{traj_optimization} into smooth commands for the manipulator, we employ an interpolation scheme. For a given time period $T$ required to move from one keyframe to the next, the commanded joint trajectory $\bm{q}_a(t)$ between the $k$th step to the $(k+1)$th step is defined as:
\begin{equation}
    \bm{q_a}(t) = (1-\frac{t}{T})\bm{q}_{a,k} +\frac{t}{T}\bm{q}_{a,k+1}
\end{equation}
In our experiments, we choose $T=10s$ using linear interpolation, sampled at 100Hz, to provide a smooth reference for a low-level controller. Upon reaching each keyframe, the manipulators remained stationary for an additional 5s to allow any residual vibrations in the hDLO to settle. Consequently, a typical 10-keyframe trajectory was completed in 150 s, ensuring that all experimental measurements reflect the steady-state equilibrium of the assembly.

\begin{figure*}
    \centering
    \includegraphics[width=\linewidth]{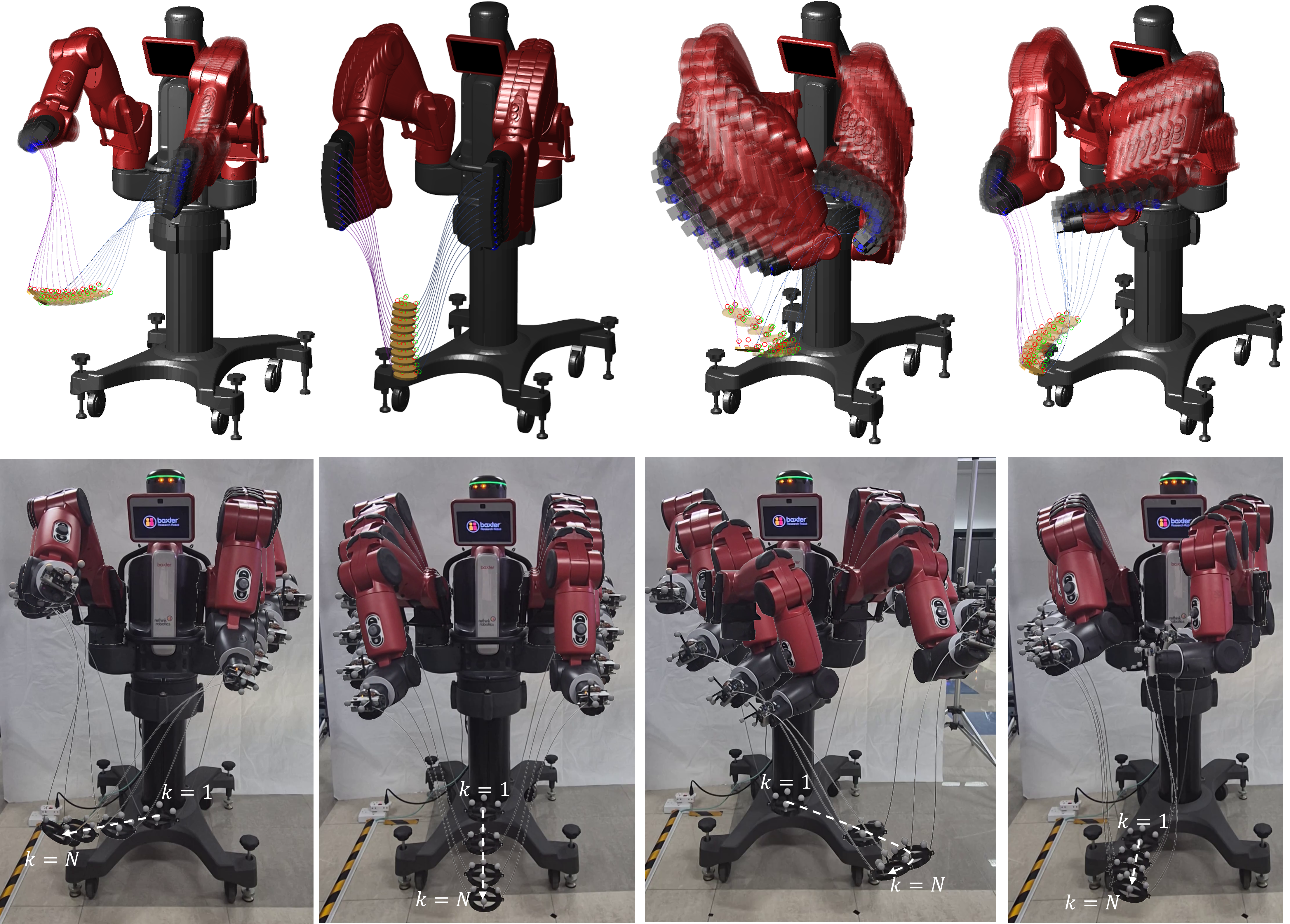}
    \caption{The unconstrained path planning is compared with the experiment. We find that the average error between the planned marker positions and those measured from the experiment is 2.6 cm or 3.7\% of the deformable object length.}
    \label{fig:Unconstrained_trajectory}
\end{figure*}
\begin{figure*}
    \centering
    \includegraphics[width=\linewidth]{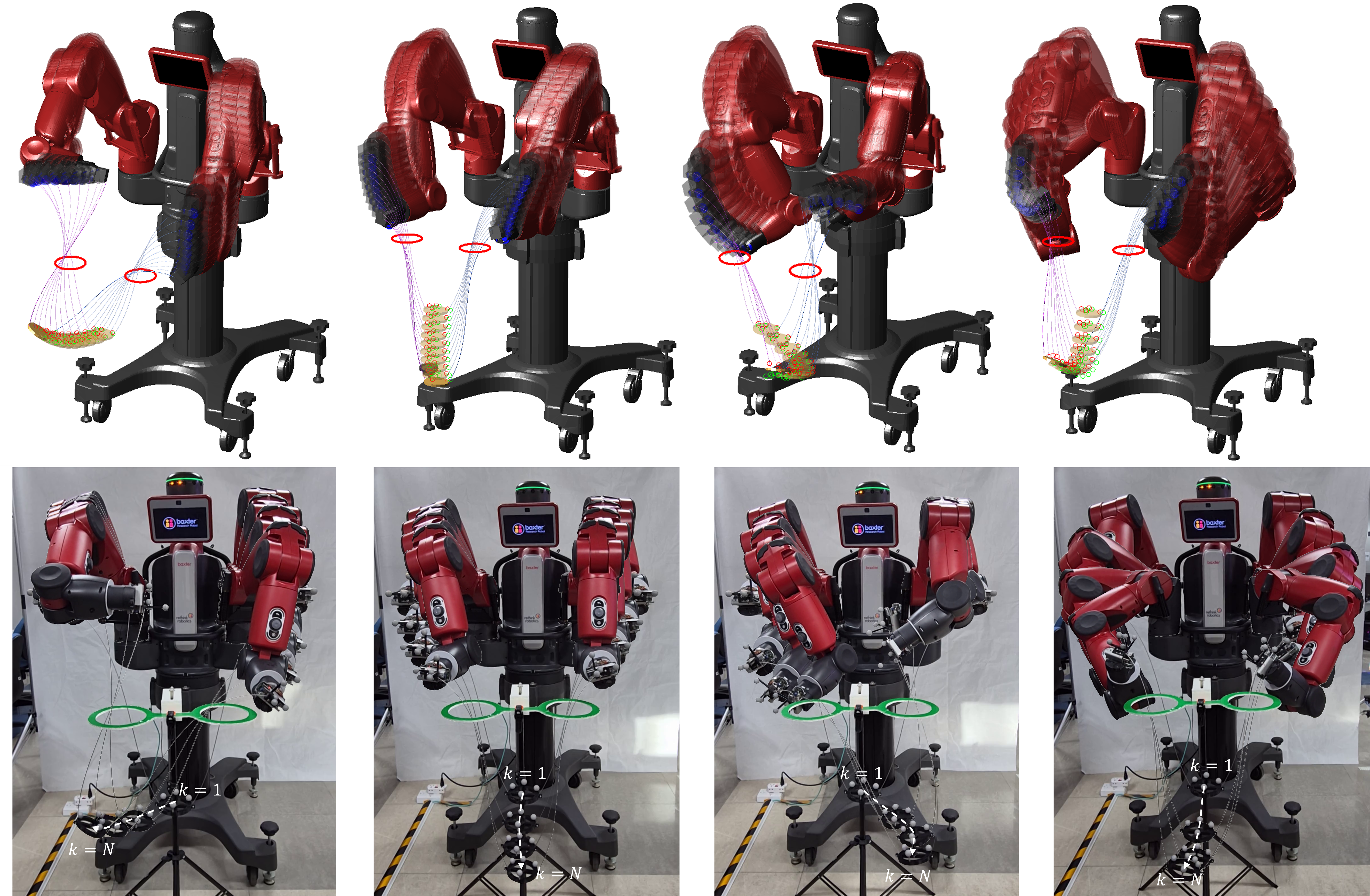}
    \caption{Constrained trajectory planning is compared with the experiment. The markers measured in the experiment are shown in green, and the markers in the simulation are shown in red. We find an average error of 0.0202m.}
    \label{fig:Trajectory_experiments}
\end{figure*}

The comparison of marker positions between the experiments and the planned path in the unconstrained scenario is shown in Fig.~\ref{fig:Unconstrained_trajectory}, while the corresponding constrained case is shown in Fig.~\ref{fig:Trajectory_experiments}. The starting and desired poses of the hDLO are the same in both cases; however, the trajectories generated by the planner differ. The average error between the planned trajectory and the achieved trajectory in terms of \eqref{error_function} is shown in Table \ref{tab:inverse_solution}. Moreover, we also show the position error between the final position of the hDLO and the desired position. To put these error values in context, the percentage error with respect to the length of one of the DLO links is also given. All simulated trajectories achieve the desired final end-effector pose within numerical precision. The small discrepancy observed in experiments lies in the range of the forward-model accuracy previously quantified, suggesting that the residual error is dominated by sim-to-real effects rather than planning inaccuracies. Although closed-loop local control could further improve tracking performance, this work concentrates on assessing the capabilities of the proposed planner.


\begin{table}[]
    \centering
    \begin{tabular}{c cc cc}
    \hline
    & \multicolumn{2}{c}{Constrained error} & \multicolumn{2}{c}{Unconstrained error} \\
    \cline{2-5}
    Trajectory & Average (cm) &Final (cm) & Avg (cm) & Final(cm) \\
    \hline
    1 & 1.37(2.01\%) & 2.62 & 1.95(2.87\%) & 1.41 \\
    2 & 1.68(2.47\%) & 2.14 & 1.64(2.41\%) & 2.16 \\
    3 & 2.18(3.20\%) & 7.53 & 3.71(5.45\%) & 4.87 \\
    4 & 2.60(3.82\%) & 3.66 & 2.82(4.14\%) & 2.78 \\
    \hline
    \end{tabular}
    \caption{Average error between the planned path and the measured path, expressed both in centimeters and as a percentage of the length of one Nitinol rod.}
    \label{tab:inverse_solution}
\end{table}

\section{Comparison with sampling based planner}\label{Comparison}
To evaluate the performance of the proposed TO planner, we implemented a bidirectional rapidly exploring random tree (BiRRT) as a sampling-based baseline. BiRRT is widely used in DLO manipulation due to its efficiency in exploring high-dimensional spaces.

In this implementation, the algorithm grows two trees, $T_{\text{start}}$ and $T_{\text{end}}$, initialised with nodes $N_\text{s}$ and $N_e$ at the initial and final configurations, respectively, until a connection is established between them. The final configuration for initialising $T_{\text{end}}$ is obtained by solving IKS \eqref{InvKinStatic}. The implementation of BiRRT is adapted for strain-based modelling of DLO by defining equilibrium conditions for the hDLO in strain-based coordinates. This ensures that the resulting trajectory is physically feasible and consistent with hDLO elastic properties, providing a direct comparison between the two methods.
Every node $N_l \in T$ store the tuple $N_l =(\bm{q}_a,\;\bm{x}_{eq})$ where $\bm{q}_a$ defines the actuated DoF that is randomly sampled and $\bm{x}_{eq} = [\bm{q}\; \bm{u}\;\bm{\lambda}]$ is the equilibrium state of the hDLO computed from $\bm{q}_a$. We define the $\bm{q}_a(N_l)$ as the projection of Node $N_l$ onto the manipulator configuration component. The shape of the deformable object can be obtained from the tuple through the recursive forward kinematics described in \eqref{fwdKinematics}. The high-level algorithm for the BiRRT implemented here is given in Algorithm \ref{alg:birrt}.

\subsection{Implementation}
The BiRRT algorithm consists of four principal steps: (1) random sampling in the actuation space, (2) extension of the search tree toward the sampled $\bm q_a$, (3) connection of the two trees, and (4) extraction of a feasible path from the initial state to the goal state through the connected tree. The specific implementation details of each step are described below.
\subsubsection{Random sampling \texttt{SampleState()}}\label{random_sample}
The joint angle space is bounded by the joint limits of the manipulator $(\bm{q}_{a,min},\bm{q}_{a,max})$. A boolean flag, \texttt{growFromStart}, is used to alternate the expansion between $T_{\text{start}}$ and $T_{\text{end}}$. At the first iteration, $T_{\text{start}}$ is selected for expansion and a random joint configuration $\bm{q}_{a}^{\text{rand}}$ is uniformly sampled within the actuator bounds. To accelerate convergence and facilitate a faster connection between $T_\text{start}$ and $T_\text{end}$, we employ a goal-biasing heuristic.

With a probability $p_{\text{goal}}=10\%$, the algorithm bypasses uniform sampling and instead sets $\bm{q}_a^{\text{rand}}$ to be the root of the opposite tree. This strategy creates a directed expansion between the two trees, significantly reducing the exploration of irrelevant regions of the high-dimensional joint space and promoting a more efficient bridge between the initial and target hDLO states.

\subsubsection{Extending the tree \texttt{Extend()}}\label{Extend}
Given the sampled joint configuration $\bm{q}_{a}^\text{rand}$, the closest node $N^*$ in the selected tree is identified using the joint space distance metric.
\begin{equation}
    N^* = \arg\min_{N_l \in T} \| \bm{q}_{a}(N_l) - \bm{q}_{a}^\mathrm{rand} \|
\end{equation}
A small step of $\gamma$ is taken from $\bm{q}_{a}(N^*)$ towards $\bm{q}_{a}^\text{rand}$ in the manipulator joint space to generate a candidate configuration, $\mathbf{q}_a^{\text{cand}}$.
\begin{equation}
\mathbf{q}_a^{\text{cand}}=\mathbf{q}_a(N_j)+\gamma\frac{\mathbf{q}_a^{\text{rand}} - \mathbf{q}_a(N^*)}{\left\|\mathbf{q}_a^{\text{rand}} - \mathbf{q}_a(N^*)\right\|}
\end{equation}
However, the candidate configuration does not, in general, satisfy the static equilibrium constraint of the deformable object and therefore requires a projection to the equilibrium manifold. The projection is carried out by solving the static equilibrium equations \eqref{statics1} and \eqref{statics2} using a root-finding procedure. To reduce computational cost, the equilibrium state associated with the nearest node $\bm{x}_{eq}(N^*)$ is used as an initial guess for the solver.

Once the projected equilibrium is obtained, feasibility along the edge connecting $N^*$ and the candidate node is verified. Specifically, the strain parameters of the deformable object are linearly interpolated between the two configurations, and the resulting intermediate states are checked against environmental and task-specific constraints. For computational efficiency, feasibility is evaluated at a finite number of interpolation points along this segment. If all checks are satisfied, the candidate node is added to the tree.

\subsubsection{Tree connection \texttt{Connect()}}\label{connect}
Whenever a new node is successfully added to one tree, the algorithm attempts to connect the opposite tree to this node using a greedy extension strategy. The closest node in the opposite tree is repeatedly extended toward the new node using the same constrained extension procedure described above, until either the two trees meet within a predefined threshold or an infeasible configuration is encountered. If infeasibility is detected during this process, the connection attempt is terminated, and the algorithm proceeds to the next iteration by sampling a new random configuration. The planning process continues until a valid connection between the two trees is established. The path is extracted with a tree search algorithm over the connected tree, yielding a feasible path from the initial configuration to the goal configuration (\texttt{ExtractPath()}).

\begin{algorithm}
\caption{Bi-directional RRT for Deformable Object Manipulation}
\label{alg:birrt}
\begin{algorithmic}[1]
\Require 
$N_s = (\bm{q}_{a,s}, \bm{x}_{eq,s})$, 
$N_e = (\bm{q}_{a,e}, \bm{x}_{eq,e})$, \\
\quad maximum iterations $N_{\max}$, 
step size $\gamma$, 
goal bias $p_{\mathrm{goal}}$
\Ensure Path $\mathcal{P}$ from $N_s$ to $N_e$ or \texttt{failure}

\State $T_{\text{start}} \leftarrow \{N_s\}$ 
\State $T_{\text{end}} \leftarrow \{N_e\}$ 
\State $\texttt{growFromStart} \leftarrow \texttt{true}$

\For{$k = 1$ to $N_{\max}$}

    \If{$\operatorname{rand}() < p_{\mathrm{goal}}$}
        \State $N_{\mathrm{rand}} \leftarrow 
        \begin{cases}
        N_e & \text{if } \texttt{growFromStart} \\
        N_s & \text{otherwise}
        \end{cases}$
    \Else
        \State $N_{\mathrm{rand}} \leftarrow \texttt{SampleState}()$
    \EndIf

    \State $(T_{\mathrm{grow}}, T_{\mathrm{other}}) \leftarrow
    \begin{cases}
    (T_{\text{start}}, T_{\text{end}}) & \text{if } \texttt{growFromStart} \\
    (T_{\text{end}}, T_{\text{start}}) & \text{otherwise}
    \end{cases}$

    \State $N_{\mathrm{new}} \leftarrow 
    \texttt{Extend}(T_{\mathrm{grow}}, N_{\mathrm{rand}}, \gamma)$

    \If{$N_{\mathrm{new}} \neq \varnothing$}
        \State $N_{\mathrm{conn}} \leftarrow
        \texttt{Connect}(T_{\mathrm{other}}, N_{\mathrm{new}}, \gamma)$
        \If{$N_{\mathrm{conn}} \neq \varnothing$}
            \State $\mathcal{P} \leftarrow
            \texttt{ExtractPath}(T_{\text{start}}, T_{\text{end}}, 
            N_{\mathrm{new}}, N_{\mathrm{conn}})$
            \State \Return $\mathcal{P}$
        \EndIf
    \EndIf

    \State $\texttt{growFromStart} \leftarrow \textbf{not }\texttt{growFromStart}$

\EndFor

\State \Return \texttt{failure}
\end{algorithmic}
\end{algorithm}

\subsection{Results}
We apply the BiRRT planner described above to the constrained environment to achieve the same hDLO end-effector poses as in Fig \ref{fig:Trajectory_experiments}. The main difference between the proposed TO method and the sampling-based method is in the computation of the new step. The TO computes a deterministic, gradient-based step that locally improves the feasibility and the optimality of the solution, whereas RRT generates a stochastic extension to explore the global configuration space. The computation time of the proposed TO planner and the BiRRT planner is compared in Table~\ref{tab:Convergence_time}. We observe that the proposed TO method can compute feasible and optimal trajectories in a time comparable to that of the sampling-based approach. Notably, for tasks requiring significant out-of-plane deformation, the TO planner demonstrates superior performance. In the case of trajectory 3, it achieves a solution nearly an order of magnitude faster.

\begin{figure}
    \centering
    \includegraphics[width=\linewidth]{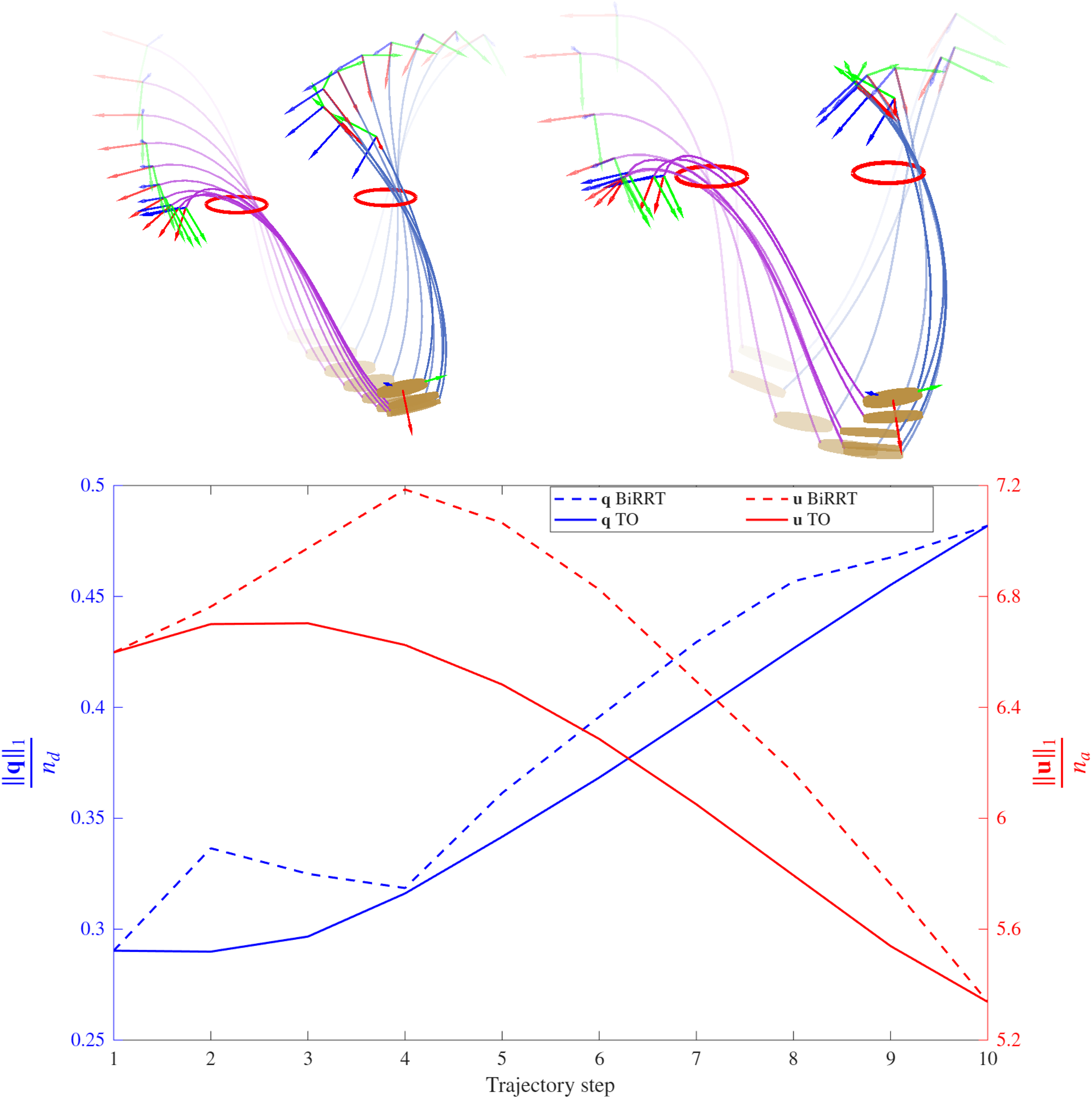}
    \caption{The trajectories computed from the TO (left) and the BiRRT (right) are compared. The TO approach finds a smoother trajectory in generalized coordinates and actuation forces.}
    \label{fig:Sampling_optim_comparison}
\end{figure}

Beyond computational efficiency, the optimization-based method offers additional behavioural advantages. Specifically, it generates trajectories that are optimal with respect to a user-defined cost function, enabling systematic trade-offs between actuation effort and motion characteristics. In Fig. \ref{fig:Sampling_optim_comparison}, we compare the performance of both planners for trajectory 3 by examining the average absolute values of joint coordinates and actuation inputs over the trajectory. The smoother profile produced by the TO planner contributes to better adherence to the quasi-static assumptions underlying both planning methods.

\begin{table}[]
    \centering
    \begin{tabular}{c c c}
    \hline
        Trajectory number & BiRRT & TO \\\hline
        1 & 33.56 (4.52) & 42.36\\
        2 & 12.34 (4.50) & 40.00\\
        3 & 201.69 (14.39) & 45.65\\
        4 & 39.77 (2.96) & 33.64\\ \hline
    \end{tabular}
    \caption{Time to convergence between our TO method and BiRRT method inspired by the literature. To account for the stochastic nature of BiRRT, each trajectory is run 5 times to get the mean and standard deviation (in brackets) of the computation time.}
    \label{tab:Convergence_time}
\end{table}

\section{Conclusion}\label{Conclusion}
This work presents a gradient-based trajectory optimization framework for manipulating hybrid deformable linear objects in constrained environments. Leveraging the GVS formulation, deformable and rigid components are modelled within a unified mathematical framework, enabling consistent kinematic and static analysis of complex assemblies. The planning problem is formulated as a quasi-static trajectory optimization, and analytical derivatives of the GVS model are supplied to the solver, making it computationally tractable.

Simulation studies on multiple hDLO assemblies and experiments on a representative hDLO demonstrate that the proposed method can generate feasible manipulation trajectories under narrow environmental constraints. Real-world experiments with a dual-arm robot validate the physical accuracy of the model, with an average discrepancy of approximately $2\;\mathrm{cm}$ between simulated and measured end-effector poses. These results show that the planner can exploit structural compliance to maneuver through confined spaces while maintaining accurate control of the hDLO end-effector pose. We also compare the proposed TO planner with a sampling-based planner adapted to strain-based modeling, highlighting differences in the resulting trajectories produced by the two approaches.

However, several aspects of the research warrant further investigation. The current framework operates under an open-loop control regime and is therefore highly sensitive to the identified parameters. Moreover, it assumes a quasi-static trajectory, which limits performance in uncertain conditions and during dynamic motions. Further work will extend the approach towards dynamic manipulation, incorporating contact dynamics so that contacts can be leveraged to achieve a larger set of feasible hDLO end-effector poses in the constrained environment, and providing feedback control for tracking the generated trajectory. 
This study shows that combining strain-based modelling of hDLOs with differentiable optimizations provides a powerful tool for manipulation planning in constrained environments. We believe this framework represents an important step toward reliable manipulation of complex deformable systems, offering new opportunities for the robotics community to tackle increasingly constrained and contact-rich environments through gradient-based planning.

\appendices
\section{Preliminary definitions}
The strain and velocity twists in \eqref{kinematics} are defined as:
\begin{align}
    \hat{\bm{\xi}_i} = \begin{bmatrix}\tilde{\bm{k}_i} & \bm{p}_i\\\mathbf{0} &0\end{bmatrix} \in \mathfrak{se}(3)\\
    \hat{\bm{\eta}_i} = \begin{bmatrix}
        \tilde{\bm{\omega}} & \bm{v}\\\mathbf{0} & 0
    \end{bmatrix} \in \mathfrak{se}(3)
\end{align}
where $\hat{\bm{\xi}_i}$ and $\hat{\bm{\eta}_i}$ are the strain and velocity twists represented in $\mathfrak{se}(3)$.
Where $\bm{k}_i \in \mathbb{R}^3$ and $\bm{p}_i\in \mathbb{R}^3$ are the local angular and linear strains at $X$, respectively. $\bm{k}_i$ and $\bm{\omega}_i$ can be represented by the corresponding skew-symmetric matrix in $\mathfrak{so}(3)$ denoted by $\tilde{\bm{k}_i}$ and $\tilde{\bm{\omega}_i}$.

\section{Analytical derivatives for planning}\label{appendix1}
\renewcommand{\thesubsection}{\thesection-\Roman{subsection}}
\subsection{Inverse kinetostatics}\label{Ap:InvKinematics_derivatives}
The analytical derivatives of the objective function of the IKS \eqref{InvKinStatic} problem are given by:
\begin{equation}\label{Jacobian_IKS_objective}
    \bm{J}_o = \begin{bmatrix}
        \frac{\partial\frac{1}{2}\|\log(\bm{g}_r^{-1}\bm{g}_l(\bm{q}))^\vee\|^2}{\partial\bm{q}}\\
        \bm{0}_{n_a\times 1}\\
        \bm{0}_{n_c\times 1}\\
        \bm{0}_{2\times 1}
    \end{bmatrix}
\end{equation}

To find the derivative of the logmap we define the distance between two transformation matrices $\bm{g}_1$ and $\bm{g}_2$ is given by the logmap as
\begin{equation}
    \bm{\epsilon} = \mathrm{log}\left(\bm{g}_1^{-1}\bm{g}_2\right)^\vee
\end{equation}
To find the derivative, we can rewrite the expression as:
\begin{equation}
    \exp(\bm{\epsilon}) = \bm{g}_1^{-1}\bm{g}_2
\end{equation}
from \cite{mathew_analytical_2025}, we know that
\begin{equation}\label{temp}
    \frac{\partial \exp(\bm{\epsilon})}{\partial q_k} = \exp({\bm{\epsilon}})\left(Ad_{\exp(\bm{\epsilon})}^{-1}{\bm{T}}(\bm{\epsilon})\frac{\partial \bm{\epsilon}}{\partial {q_k}}\right)^\wedge
\end{equation}
where $\bm{T}()$ is the tangent operator of the exponential map given by:
\begin{equation}
    \bm{T}(\bm{\Omega}) =\int_0^1\exp\left(s\cdot ad_{\bm{\Omega}}\right)ds
\end{equation}

\eqref{temp} can be rewritten using the geometric Jacobian as:
\begin{equation}
    \exp(\bm{\epsilon}){\left(Ad_{\exp(\bm{\epsilon})}^{-1}\bm{T}(\bm{\epsilon})\frac{\partial\bm{\epsilon}}{\partial {q_k}}\right)^\wedge} =\exp(\bm{\epsilon})\hat{\bm{S}_k}
\end{equation}
where $\bm{S}_k$ is the $k$th column of the relative jacobian betweeen 1 and 2. Extending to $\frac{\partial}{\partial\bm{q}}$ and rearranging the terms, we get:
\begin{equation}
    \frac{\partial\bm{\epsilon}}{\partial\bm{q}}=\bm{T}(\bm{\epsilon})^{-1}\left(\mathrm{Ad}_{exp(\bm{\epsilon})}\bm{J}_2-\bm{J}_1\right)\label{twist_wrt_q}
\end{equation}
Our close chain constraint is defined through this derivative where $\bm{A} =\frac{\partial \bm{\epsilon}}{\partial \bm{q}}$. So we have a constraining force given by $\bm{A}^T\bm{\lambda}$. However, in our implementation, we define a term $$\bar{\bm{A}} = \left(\mathrm{Ad}_{exp(\epsilon)}\bm{J}_2-\bm{J}_1\right)$$
Our solution will still remain the same because the constraining force will be defined by $\bar{\bm{A}}^T\bar{\bm{\lambda}}$ where $\bar{\bm{\lambda}} =(\bm{T}(\epsilon)^{-1})^T\bm{\lambda}$. Since $\bar{\bm{\lambda}}$ is obtained through an optimization and not computed explicitly, this modification avoids unnecessary tangent operator computation. 

In the case of the $\bm{g}_1$ being constant as in the objective function, we have $\bm{J}_1=\bm{0}$ to give:
\begin{equation}
    \frac{\partial\bm{\epsilon}}{\partial\bm{q}}=\bm{T}(\epsilon)^{-1}\mathrm{Ad}_{exp(\epsilon)}\bm{J}_2
\end{equation}

The analytical derivatives with respect to the other optimization variables are zeros.

The analytical jacobian of the equality and inequality constraints of \eqref{InvKinStatic} are given by: 
\begin{equation}
\resizebox{0.85\columnwidth}{!}{$
    \bm{J}_{ce} = \begin{bmatrix}
        \bm{K} - \frac{\partial \bm{B}}{\partial \bm{q}}\bm{u}-\frac{\partial\bm{F}}{\partial \bm{q}} - \frac{\partial\bm{A}^T}{\partial\bm{q}}\bm{\lambda} &-\bm{B} & -\bm{A}^T & 0\\
        \bm{A}^T&\bm{0}_{6\times n_a} & \bm{0}_{6\times n_c} & 0\\
         \frac{\partial c_{eq}(\bm{q},\bm{X}^\dagger)}{\partial\bm{q}} & \bm{0}_{1\times n_a} & \bm{0}_{1\times n_c} & \frac{\partial c_{eq}(\bm{q},\bm{X}^\dagger)}{\partial \bm{X}^\dagger}\\   
    \end{bmatrix}
$}
\end{equation}
\begin{equation}\label{constraints_jacobian}
\bm{J}_{cin} = 
\begin{bmatrix}
    \frac{\partial c_{in}(\bm{q},\bm{X}^\dagger)}{\partial\bm{q}} & \bm{0}_{1\times n_a} & \bm{0}_{1\times n_c} & \frac{\partial c_{in}(\bm{q},\bm{X}^\dagger)}{\partial \bm{X}^\dagger}
    \end{bmatrix}
\end{equation}
The analytical terms for $\frac{\partial \bm{B}}{\partial \bm{q}}$, $\frac{\partial \bm{F}}{\partial \bm{q}}$ and $\frac{\partial \bm{A}^T}{\partial \bm{q}}$ are derived in \cite{mathew_analytical_2025}. The derivatives for the environmental constraints are given by:
\begin{align}\label{env_constraint_derivatives}
    \frac{\partial c_{eq}}{\partial \bm{q}} =& -\frac{\partial z(\bm{q},X^\dagger)}{\partial \bm{q}}\\
    \frac{\partial c_{in}}{\partial\bm{q}} =& -2\left(\bm{P}_h-\bm{P}(\bm{q},X^\dagger) \right)\frac{\partial \bm{P}(\bm{q}, X^\dagger)}{\partial \bm{q}}
\end{align}
But $\bm{P}(\bm{q},X^\dagger) = [x(\bm{q},X^\dagger)\; y(\bm{q},X^\dagger)]$, so to find the analytical derivatives of the environmental constraints, we need the derivatives of the position, $\frac{\partial\bm{r}(\bm{q},X^\dagger)}{\partial \bm{q}}$ and $\frac{\partial\bm{r}(\bm{q},X^\dagger)}{\partial X^\dagger}$
\subsection{Position derivatives}\label{Ap:postion_derivatives}
The derivatives of $c_{eq}$ and $c_{in}$ with respect to the optimization variables depend on the derivative of the position of the DLO link in the hDLO. The pose of the DLO cross section at an arbitrary length $X^\dagger$ is found through the interpolation of known discrete poses along the length as:
\begin{equation}
    \bm{g}(X^\dagger) = \bm{g}_j\bm{g}_\alpha
\end{equation}
where $\bm{g}_\alpha = \exp\left(\left(\frac{X^{\dagger} - X_j}{X_{j+1}-X_{j}}\right)\bm{\Omega_j})\right)$. From this, we can say that $\bm{r}(X^\dagger) = \bm{R}_j\bm{r}_\alpha +\bm{r}_j$. differentiating this with respect to the $k$th component $q_k$ of the vector $\bm{q}$, we get:
\begin{align}
    \frac{\partial \bm{r}(X^{\dagger})}{\partial q_k} &= \frac{\partial(\bm{R}_j \bm{r}_\alpha)}{\partial q_k} +\frac{\partial \bm{r}_j}{\partial q_k}\\
    &=\frac{\partial \bm{R}_j}{\partial q_k}\bm{r}_\alpha +\bm{R}_j\frac{\partial \bm{r}_\alpha}{\partial q_k} +\frac{\partial \bm{r}_j}{\partial q_k}
    \label{position_derivative}
\end{align}
We find each of these terms separately. Using \eqref{kinematics}, we have:
\begin{align}
    \dot{\bm{R}} &= \bm{R}\tilde{\bm{\omega}}\\
    \dot{\bm{r}} &= \bm{Rv}
\end{align}
The partial derivative with respect to the $k$th component of $\bm{q}$ will then be:
\begin{align}
    \frac{\partial\bm{R}_j}{\partial q_k}\bm{r}_\alpha =& \bm{R}_j \tilde{\bm{J}}_{j_k}^\omega\bm{r}_\alpha =-\bm{R}_j \tilde{\bm{r}}_\alpha\bm{J}_{j_k}^\omega\\\
    \frac{\partial\bm{r}_j}{\partial q_k} =& \bm{R}_j \bm{J}_{j_k}^v
\end{align}
Combining all the columns of $\bm{q}$, we get:
\begin{align}
    \frac{\partial\bm{R}_j}{\partial \bm{q}}\bm{r}_\alpha =&-\bm{R}_j \tilde{\bm{r}}_\alpha\bm{J}_j^{\omega}\\\
    \frac{\partial\bm{r}_j}{\partial \bm{q}} =& \bm{R}_j \bm{J}_j^{v}\label{position_wrt_q}
\end{align}

$\frac{\bm{r}_\alpha}{\partial \bm{q}}$ follows the same structure as \eqref{position_wrt_q}:
\begin{equation}
    \frac{\partial \bm{r}_\alpha}{\partial\bm{q}} = \bm{R}_\alpha \begin{bmatrix}\bm{0} & \bm{I}\end{bmatrix}\text{Ad}^{-1}_{\exp{(\alpha \hat{\Omega}})} \bm{T}(\alpha\Omega)\alpha \frac{\partial \Omega}{\partial \bm{q}}
\end{equation}
using \eqref{twist_wrt_q}, we have
\begin{equation}
      \frac{\partial \Omega}{\partial \bm{q}} = \bm{T}^{-1}(\bm{\Omega}_j)\left(\text{Ad}_{\exp{( \hat{\Omega}})}\bm{J}_{j+1} - \bm{J}_{{j}}\right)
\end{equation}
with

The analytical derivatives of the position of the deforamble object at $X^{\dagger}$ with respect to $\bm{u}$ and $\bm{\lambda}$ are zeros. However, we need to find the derivative with respect to $X^{\dagger}$. Between two computational points $X_j$ and $X_{j+1}$, $\bm{R}_j$ and $\bm{r}_j$ are constant, so the positional derivative with respect to $X$ is given by:
\begin{equation}
    \frac{\partial \bm{r}(X^{\dagger})}{\partial X^{\dagger}} = \bm{R}_j\frac{\partial \bm{r}_\alpha}{\partial X^{\dagger}}+\bm{r}_j
\end{equation}

The spatial derivative is known to us through the definition of strain \eqref{spacial_derivative}. 
\begin{equation}
     \frac{\partial \bm{r}(X^\dagger)}{\partial X^{\dagger}} = \bm{g}_j\bm{g}_\alpha\bm{\Omega}_j\begin{bmatrix}
         \bm{0}_{3\times 1}\\1
     \end{bmatrix}
\end{equation}

The analytical derivatives of the constraints in \eqref{constraints_jacobian} can then be found by substituting position derivatives in \eqref{env_constraint_derivatives}.

\subsection{Analytical derivatives for the trajectory generation}\label{Ap:path_planning_derivatives}

The analytical derivatives of the objective function and the constraints of the TO problem \eqref{traj_optimization} are given by:
\begin{equation}
\begin{split}
    \bm{J}_o &= \frac{\partial f_p(\bm{x})}{\partial\bm{x}}\\
    \end{split}
\end{equation}
The path cost function \eqref{terminal_cost_1} used for finding a smooth trajectory between the initial state and the final state are given by:
\begin{align}
    f_p &= \sum_{k=1}^{N-1}\|\bm{q}_{k+1}-\bm{q}_k\|^2_{Q_q}+\|\bm{u}_{k+1}-\bm{u}_k\|^2_{Q_u}+\notag\\&\qquad\|\bm{\lambda}_{k+1}-\bm{\lambda}_k\|^2_{Q_\lambda}
\end{align}
The optimization variables are $\bm{q}$,$\bm{u}$, $\bm{\lambda}$ and $X$ at timesteps $k=1,2,3 ...N$. When $k=1$, it appears twice in the series. So the derivative is 
\begin{equation}
   \frac{\partial f_p}{\partial \bm{q}_2} = 2\bm{Q}_q(\bm{q}_2-\bm{q}_1)-2\bm{Q}_q(\bm{q}_3-\bm{q}_2)
\end{equation}
And similarly for all values of $k\neq N$. At $k=N$, $q_N$ appears once, and the derivative is given by:
\begin{equation}
    \frac{\partial f_p}{\partial \bm{q}_N} = 2\bm{Q}_q(\bm{q}_N-\bm{q}_{N-1})
\end{equation}

The gradient with respect to $\bm{q}$ is given by:
\begin{align}
\frac{\partial f_p}{\partial \bm{q}_k} &= 
\begin{cases}
2\,\bm{Q}_q(\bm{q}_k - \bm{q}_{k-1}) - 2\,\bm{Q}_q(\bm{q}_{k+1} - \bm{q}_{k}), & 1\leq k<N, \\[6pt]
2\,\bm{Q}_q(\bm{q}_k - \bm{q}_{k-1}), & k=N.
\end{cases}
\end{align}

The derivatives with respect to $\bm{u}$ and $\bm{\lambda}$ have a similar structure.


The derivative of the constraints, $\bm{J}_t$, will take a block-diagonal form with the derivatives of the IKS constraints filling the diagonal positions, since each constraint depends only on the variables at the corresponding keyframe. i.e. there is no coupling between keyframes. The analytical derivative will be of the form
\begin{equation}
    \bm{J}_t = \begin{bmatrix}
        \bm{J}_{ce1} & 0 &\cdots& 0\\
        0&\bm{J}_{ce2} & \cdots& 0\\
        \vdots & \vdots & \ddots & \vdots\\
        0 &0 & \cdots & \bm{J}_{ceN}
    \end{bmatrix}
\end{equation}

\section*{Acknowledgments}

Omitted for anonymous review

\ifCLASSOPTIONcaptionsoff
  \newpage
\fi

\bibliographystyle{IEEEtran}
\bibliography{ref}


\end{document}